\documentclass{SCIS2026}

\usepackage[T1]{fontenc}
\usepackage{lmodern}
\usepackage{makecell}
\usepackage{fix-cm}
\usepackage{bbm}
\usepackage{multirow}
\usepackage{xspace}
\newcommand{\ie}{{\emph{i.e.}},\xspace}

\usepackage{bbding}

\newcommand{\tablesmallfont}{\fontsize{6}{12}\selectfont }

\begin{document}

\ArticleType{RESEARCH PAPER}

\Year{2025}
\Month{January}
\Vol{68}
\No{1}
\DOI{}
\ArtNo{}
\ReceiveDate{}
\ReviseDate{}
\AcceptDate{}
\OnlineDate{}
\AuthorMark{}
\AuthorCitation{}

\title{Structured prototype regularization for synthetic-to-real driving scene parsing}{}

\author[1]{Jiahe Fan}{}
\author[2]{Xiao Ma}{}
\author[3]{Sergey Vityazev}{}
\author[4]{George Giakos}{}
\author[1]{Shaolong Shu}{}
\author[1]{Rui Fan}{{rui.fan@ieee.org}}

\address[1]{Tongji University, Shanghai 201804, China}
\address[2]{Beijing Institute of Aerospace Control Devices, Beijing 100039, China}
\address[3]{Ryazan State Radio Engineering University, Ryazan 390005, Russia}
\address[4]{Manhattan University, New York 10471, USA}

\abstract{Driving scene parsing is critical for autonomous vehicles to operate reliably in complex real-world traffic environments. To reduce the reliance on costly pixel-level annotations, synthetic datasets with automatically generated labels have become a popular alternative. However, models trained on synthetic data often perform poorly when applied to real-world scenes due to the synthetic-to-real domain gap. Despite the success of unsupervised domain adaptation in narrowing this gap, most existing methods mainly focus on global feature alignment while overlooking the semantic structure of the feature space. As a result, semantic relations among classes are insufficiently modeled, limiting the model's ability to generalize. To address these challenges, this study introduces a novel unsupervised domain adaptation framework that explicitly regularizes semantic feature structures to significantly enhance driving scene parsing performance in real-world scenarios. Specifically, the proposed method enforces inter-class separation and intra-class compactness by leveraging class-specific prototypes, thereby enhancing the discriminability and structural coherence of feature clusters. An entropy-based noise filtering strategy improves the reliability of pseudo labels, while a pixel-level attention mechanism further refines feature alignment. Extensive experiments on representative benchmarks demonstrate that the proposed method consistently outperforms recent state-of-the-art methods. These results underscore the importance of preserving semantic structure for robust synthetic-to-real adaptation in driving scene parsing tasks.}


\maketitle

\section{Introduction}
\label{sec.introduction}

\subsection{Background}

Driving scene parsing analyzes traffic environments to identify and categorize key elements, such as roads, vehicles, traffic lights, and sidewalks~\cite{huang2024roadformer+,xue2024visual,guan2025lidar}. It is a key component of autonomous driving perception systems and provides a comprehensive machine-level understanding of traffic environments, supporting safe navigation and decision-making~\cite{chen2019autonomous,ruan2025skpnet}. Recent advances in fully supervised methods, primarily based on convolutional neural networks (CNNs), have achieved notable success by leveraging large-scale, manually annotated datasets~\cite{fan2021learning, guo2025lix,tang2025ticoss}. Nevertheless, generating such pixel-level annotations remains prohibitively costly. For instance, labeling a single image in the Cityscapes dataset~\cite{Cordts2016} requires approximately 1.5 hours under normal conditions, and up to 3.3 hours under adverse weather conditions~\cite{acdc_data}. 

\begin{figure}[t!]
    \centering
    \includegraphics[width=0.95\textwidth]{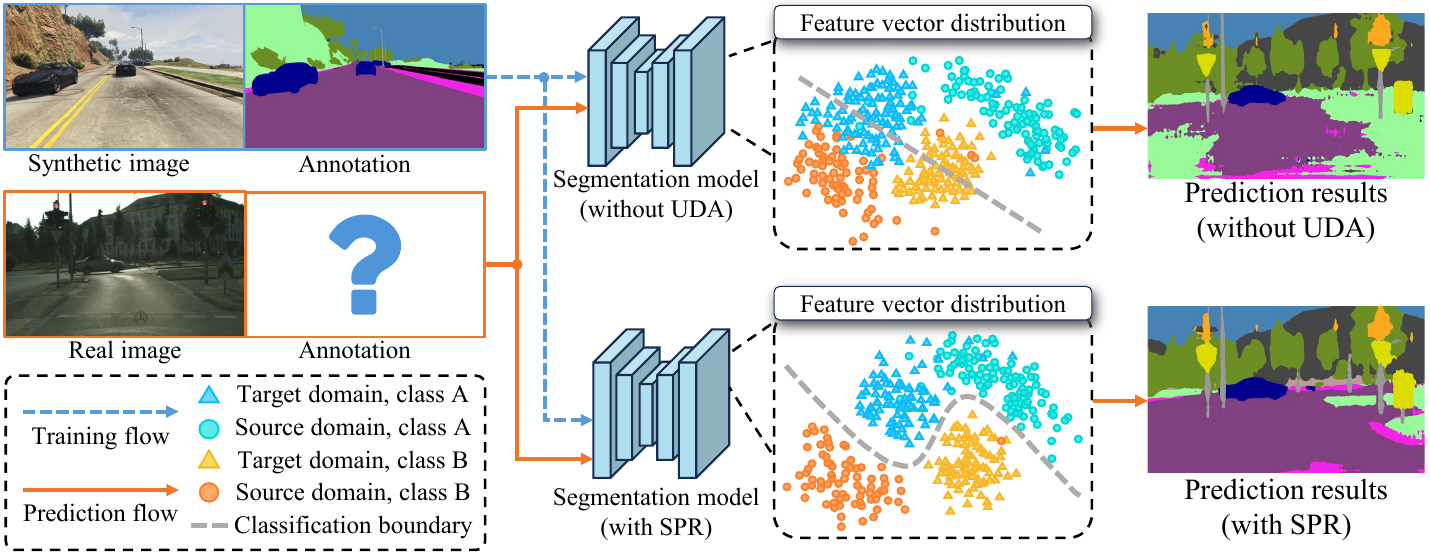}
    \caption{Image domain discrepancy and the effectiveness of the UDA method in driving scene parsing. The proposed SPR enhances feature alignment by explicitly modeling and preserving inter-class separability and intra-class compactness, leading to more structured and discriminative feature representations across domains.}
    \label{fig.illu_uda}
\end{figure}

An alternative to manual annotation is the use of synthetic datasets with automatic pixel-level labels for CNN training~\cite{Cordts2016,Ros2016}. However, models trained on synthetic data often suffer substantial performance drops when applied to real-world scenes due to the domain gap between the synthetic (source) and real (target) domains~\cite{Marsden2022}, as illustrated in Fig.~\ref{fig.illu_uda}. This discrepancy arises from differences in image characteristics such as lighting, texture, and contrast, all of which hinder the generalization of CNNs to real-world driving environments. To address this issue, unsupervised domain adaptation (UDA) methods have been developed for driving scene parsing~\cite{sun2023enhancing,zhai2024maximizing,ma2020preserving,li2019challenges,zheng_ruler_2025,li_adagpar_2025}. UDA methods transfer knowledge from labeled synthetic datasets to unlabeled real-world images by aligning the feature distributions across domains. By narrowing the gap between source and target representations, UDA provides a practical and scalable solution for enhancing model performance in real-world autonomous driving applications. This study aims to bridge the domain gap between synthetic and real-world data, thereby improving the performance of driving scene parsing models in real-world environments.

\subsection{Existing challenges and motivation}

Despite significant progress in UDA for driving scene parsing, several key challenges remain. A primary issue is that most existing methods rely on global feature alignment strategies, such as adversarial learning, which often overlook the semantic structure of the feature space~\cite{Marsden2022}. 
As a result, relationships among semantic classes, such as roads, vehicles, and pedestrians, are not explicitly preserved, which limits the effectiveness of feature alignment across domains.

Recent advances in prototype-based contrastive learning methods~\cite{Jiang2022} have improved class-wise feature alignment by associating features with semantic prototypes. Nonetheless, these methods typically treat prototypes as isolated points, neglecting the structural relationships within the feature space~\cite{Jiang2022}. Effective driving scene parsing requires not only accurate alignment of target features to source-domain prototypes, but also the preservation of a well-structured feature space characterized by high inter-class separability\footnote{Inter-class separability ensures that features from different semantic categories, such as roads, vehicles, and pedestrians, are projected into clearly separated regions in the feature space, thereby minimizing class confusion.} and strong intra-class compactness\footnote{Intra-class compactness requires that features from the same category remain tightly clustered, even across variations in appearance caused by lighting, occlusion, or domain shift.}~\cite{Zhang2021}. These two properties jointly contribute to more reliable decision boundaries and improved generalization across domains. However, most existing prototype-based contrastive learning methods do not explicitly enforce or model such structural constraints, thereby limiting their effectiveness in real-world domain adaptation tasks.
 
Moreover, existing strategies~\cite{Tsai2018,Wang2020a} often prioritize global feature alignment while overlooking the distribution of pixel-level features, which is essential for accurate driving scene parsing. This limitation becomes particularly pronounced under significant domain shifts, where pixel-wise predictions in the target domain are prone to uncertainty and noise~\cite{Zhang2021}. In the absence of effective mechanisms to account for pixel-level discrepancies, alignment may be distorted by unreliable predictions, ultimately degrading the performance of scene parsing models. Addressing this issue requires more granular strategies capable of suppressing noisy pixel features while enhancing their consistency with the class-level semantic structure.

\subsection{Novel contributions}

To address the aforementioned limitations, this study presents a novel UDA framework based on structured prototype regularization (SPR), which explicitly enforces inter-class and intra-class consistency within the feature space. Given the significant discrepancy in feature distributions between the source and target domains, driving scene parsing on target-domain images often contains substantial noise. To mitigate the impact of this noise on feature alignment, the proposed method first computes class-specific prototypes from CNN logits, which serve as approximate centroids for each semantic category. These prototypes effectively capture the core characteristics of each class and suppress the influence of noisy features, thereby enhancing representation robustness and discriminability. To regularize the semantic structure of the feature space, the proposed method quantifies the relationships among prototypes by computing both inter-class and intra-class correlation matrices. These matrices guide the formulation of regularization terms that promote a semantically well-structured feature space, characterized by clear inter-class separability and strong intra-class compactness. Building on these refined prototypes, a contrastive loss is introduced to align feature distributions across domains. Since driving scene parsing fundamentally requires a dense, pixel-level understanding of the environment, aligning pixel-level feature distributions is critical for achieving fine-grained parsing performance. To this end, the correlation matrix between each pixel-level feature and each class prototype is further computed and utilized to guide the optimization process. Specifically, this study introduces an entropy-based measure derived from the correlation matrix to filter noisy pixel-level predictions in the target domain, thereby improving the prototype estimation accuracy. Furthermore, a pixel-level attention mechanism is incorporated into the contrastive loss to emphasize more reliable features during alignment, further enhancing the consistency of pixel-level representations across domains.

The novel contributions of this study can be summarized as follows:
\begin{itemize}

\item A novel contrastive learning framework based on structured prototype regularization, which explicitly regularizes inter-class separability and intra-class compactness within the feature distributions, thereby improving CNN generalization to real-world driving scenes.
\item Structural modeling of inter-class and intra-class feature relations for domain adaptation, with imposed constraints that guide more effective cross-domain feature alignment.
\item Enhanced pixel-level feature alignment with entropy-based noise filtering and pixel-level attention, both incorporated into the contrastive learning process to improve the driving scene parsing accuracy on real-world data.

\end{itemize}

The remainder of this article is structured as follows: Section~\ref{sec.related_works} reviews related state-of-the-art methods. Section~\ref{sec.methodology} presents the technical details of the proposed SPR method. 
Section~\ref{sec.exp} presents both quantitative and qualitative experimental results, comparing the proposed method with other state-of-the-art UDA methods.
Section~\ref{sec.discussion} provides theoretical insights into the role of structured prototype regularization in improving domain adaptation performance. Finally, Section~\ref{sec.conclusion} concludes the article and discusses potential directions for future research.

\section{Related work}
\label{sec.related_works}

\subsection{Unsupervised domain adaptation for driving scene parsing}
\label{sec.related_works.uda_for_semantic_segmentation}

Driving scene parsing aims to assign pixel-level semantic labels to road scene images, enabling a detailed understanding of the driving environment~\cite{fan2020eccv,li2024roadformer,wu2024s3mnet,liu2023semi,wang2025overview}. UDA methods alleviate the heavy dependency of CNN-based segmentation models on annotated data, thereby facilitating more cost-effective real-world deployment.
Currently, UDA methods for driving scene parsing primarily fall into three categories: (1) style transfer~\cite{Hoffman2018}, (2) feature alignment~\cite{Hoffman2018,Hoffman2016}, and (3) self-training~\cite{Zou2018}. Inspired by recent advances in unpaired image-to-image translation~\cite{Hoffman2018}, style transfer methods transform the style of synthetic (source) images to resemble real-world (target) images~\cite{Hoffman2018}. Feature alignment methods~\cite{Tsai2018, Luo2021} reduce domain discrepancies by aligning feature distributions between the source and target domains. This is achieved either through statistical methods, such as minimizing the maximum mean discrepancy across different domains within domain-specific layers~\cite{Long2015a}, or through adversarial learning, where a domain discriminator is used to encourage domain-invariant feature extraction~\cite{Hoffman2016}. Self-training methods~\cite{Mei2020, Jiang2022, Zou2018} generate pseudo labels for unlabeled target-domain images and iteratively refine the model through retraining. Recent methods have incorporated more complex UDA frameworks. For instance, ADPL~\cite{ADPL-Dual_2023} proposes an adaptive dual-path learning strategy that alleviates visual inconsistency and enhances pseudo label quality by introducing two interactive single-domain adaptation paths, one for the source domain and one for the target domain. FREDOM~\cite{FREDOM_2023} introduces a fairness-aware UDA framework for semantic segmentation by imposing a fairness objective on class distributions and leveraging a conditional structure network with self-attention to model structural dependencies.

Adversarial learning reduces the discrepancy in global feature distributions across domains by encouraging CNNs to extract domain-invariant representations. However, aligning global distributions alone does not guarantee effective class separability in the shared feature space~\cite{Jiang2022}. To address this limitation, recent studies have incorporated class-specific information into the feature alignment process. For example, class labels are incorporated into the discriminator during adversarial learning to enable class-aware feature alignment and improve feature discriminability~\cite{Wang2020a}. 
Nevertheless, most existing methods neglect to explicitly model structural correlations within the feature space. Consequently, they struggle to ensure sufficient inter-class separability and intra-class compactness, thereby undermining the discriminative power of the learned representations in the target domain.

\subsection{Contrastive learning for driving scene parsing}
\label{sec.related_works.contrastive_learning}

Contrastive learning~\cite{Hadsell2006} learns feature representations by encouraging similarity between positive (semantically similar) image pairs and dissimilarity between negative (semantically different) image pairs. Typically, positive pairs can be derived from different augmentations of the same image that preserve semantic content, or from samples of the same class across domains. In contrast, negative pairs are drawn from semantically different images, potentially spanning different domains. This learning paradigm guides the model to cluster semantically consistent features while pushing apart unrelated ones. Another prominent self-supervised method is self-training, which iteratively refines pseudo labels to improve performance in the absence of ground truth~\cite{He2020, Chen2020}. For instance, STC~\cite{Jiang2022a} utilizes contrastive objectives to develop association embeddings for video instance segmentation. More recently, contrastive learning has also been successfully adapted to driving scene parsing tasks~\cite{Alonso2021}. In the field of UDA, contrastive learning has been employed to align image feature spaces across domains. CLST~\cite{Marsden2022} leverages contrastive objectives to achieve more refined domain-adapted image features. EHTDI~\cite{EHTDI_2022} explores target-domain decision boundaries and features via a mix-up strategy and multi-level contrastive learning. SePiCo~\cite{SePiCo_2023} employs semantic-guided pixel contrast, incorporating both centroid-aware and distribution-aware pixel contrast, to learn class-discriminative and balanced pixel representations across domains. CACP~\cite{CACP_2025} addresses domain adaptive semantic segmentation by leveraging both source-domain and target-domain prototypes to dynamically rectify pseudo labels, and applying contrastive learning on prototypes to enhance inter-class separability and reduce domain shifts. 

Inspired by these advances, this study presents a novel contrastive learning-based UDA framework that explicitly enforces the consistency of inter-class and intra-class relations within the feature space, thereby enhancing domain-invariant representation learning for driving scene parsing.

\section{Methodology}
\label{sec.methodology}
\subsection{Preliminaries}
\label{sec.methodology.preliminary}

In this article, the source domain is associated with a labeled dataset $\mathcal{D}_{s}$, which contains ${N_s}$ pairs of synthetic RGB images $\boldsymbol{X}_{s} \in \mathbb{R}^{H \times W \times 3}$ and corresponding pixel-level semantic annotations $\boldsymbol{Y}_{s} \in \mathbb{R}^{H \times W }$ with respect to $C$ semantic categories. The target domain is associated with an unlabeled dataset $\mathcal{D}_{t}$ containing ${N_t}$ real-world RGB images $\boldsymbol{X}_{t}\in \mathbb{R}^{H \times W \times 3}$ without corresponding semantic annotations.
The semantic segmentation model $\mathcal{M}_s$, which consists of a feature extractor $\mathcal{F}$ and a pixel-level classifier $\mathcal{C}$, is initially trained on the labeled source-domain data in a supervised manner by minimizing the following pixel-wise cross-entropy loss:
\begin{equation}
    \label{eq.cross_entropy_source}
	\mathcal{L}_{ce} =  - \mathcal{S}\left({{\boldsymbol{H}}_{s}} \odot \log (\sigma(\boldsymbol{L}_{s}))\right),
\end{equation}
where the operator $\mathcal{S}$ represents the summation over all elements of a given tensor or matrix, ${\boldsymbol{H}}_{s} \in \{0,1\}^{ H \times W \times C}$ denotes the one-hot encoding of $\boldsymbol{Y}_{s}$, $\odot$ represents the Hadamard product, $\sigma(\cdot)$ denotes the softmax function, and $\boldsymbol{L}_s\in\mathbb{R}^{H\times W\times C}$ denotes the logits produced by $\mathcal{M}_s$. 

Although the segmentation model trained on source-domain data can directly generate softmax-based label predictions $\mathcal{C}(\mathcal{F}(\boldsymbol{X}_t))$ for images in the target domain, the domain discrepancies between the source and target domains often result in performance degradation. Furthermore, the lack of pixel-wise annotations in the target dataset hinders the direct fine-tuning of the semantic segmentation model, further limiting its generalization performance.

\begin{figure*}[t!]
    \centering
    \includegraphics[width=0.99\textwidth]{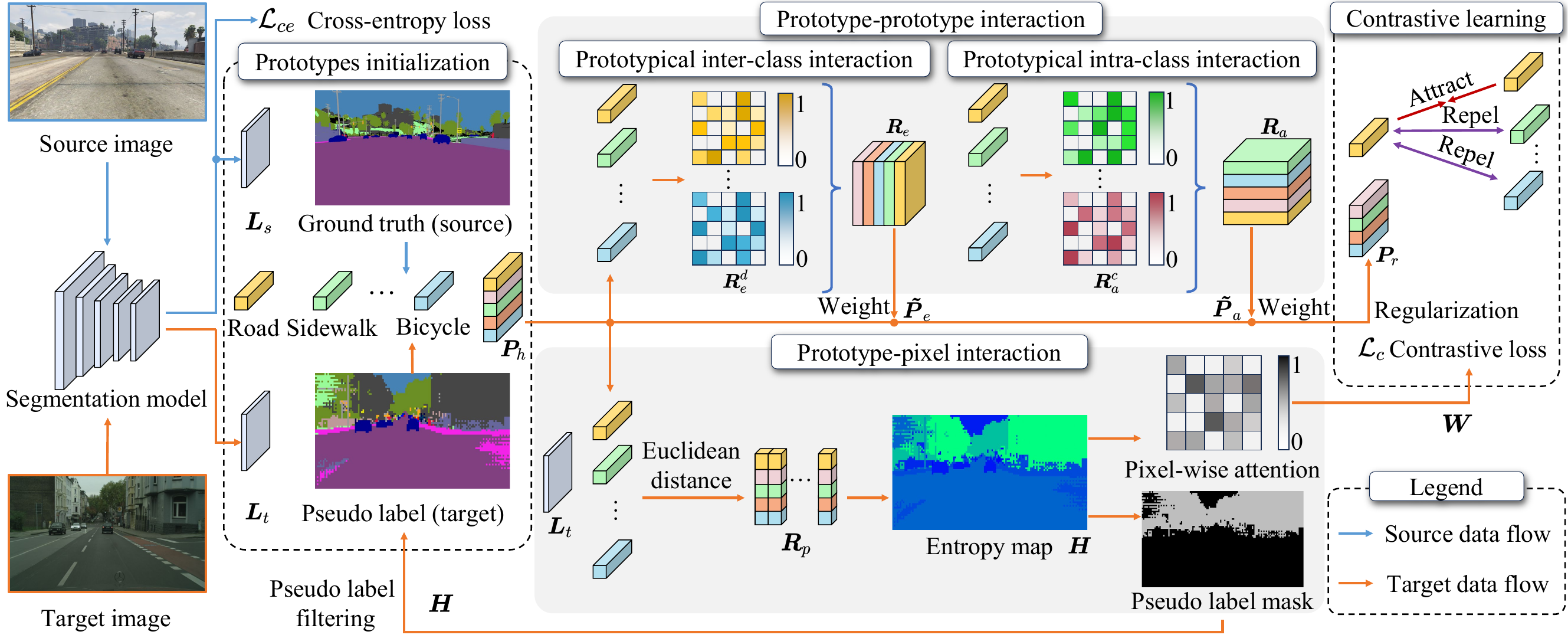}
    \caption{Overview of the proposed SPR framework. The segmentation model is trained with cross-entropy loss $\mathcal{L}_{ce}$ on the source-domain dataset and contrastive loss $\mathcal{L}_{c}$ on datasets from both domains. Prototype–prototype interactions enforce inter-class separability and intra-class compactness by structurally regularizing class prototypes. Prototype–pixel interactions align pixel-wise features with these refined prototypes, incorporating entropy-based filtering and attention mechanisms to enhance semantic consistency in the target domain.}
    \label{fig.framework}
\end{figure*}

\subsection{Prototype-prototype interaction}
\label{sec.methodology.PtoP}
\subsubsection{Class prototype estimation}
\label{sec.methodology.PtoP.initial}

Following the training of the segmentation model on the source-domain dataset, the prototype vector $\boldsymbol{p}^{c} = \left( p_{1}^c, \dots, p_{D}^c \right)^\top\in\mathbb{R}^{D}$ is defined as the centroid of the feature vectors assigned to the $c$-th class, where $D$ denotes the number of feature channels. Specifically, each element of the prototype is computed as:
\begin{equation}
    \label{eq.initial_proto}
    p_{d}^c = \frac{\mathcal{S} \big({\boldsymbol{L}_d \odot \mathbbm{1}[\boldsymbol{Y} = c]\big)}   }{\mathcal{S} \big(\mathbbm{1}[\boldsymbol{Y} = c]\big)},
\end{equation}
where $\boldsymbol{L}_{d} \in \mathbb{R}^{H \times W}$ denotes the $d$-th channel of the logits $\boldsymbol{L} \in \mathbb{R}^{H \times W \times D}$, and $\boldsymbol{Y} \in \mathbb{R}^{H \times W}$ represents the corresponding labels. 
The indicator function $\mathbbm{1}[\cdot]$ selects pixels assigned to the $c$-th class based on the labels, where $\mathbbm{1}[\boldsymbol{Y} = c] \in \{0,1\}^{H \times W}$. 
Finally, the prototype matrix $\boldsymbol{P} = \left({\boldsymbol{p}^1},{\boldsymbol{p}^2},\ldots, {\boldsymbol{p}^C}\right) \in \mathbb{R}^{D \times C}$ is constructed by concatenating the class-specific prototypes of all $C$ semantic classes, serving as an approximation of the semantic centroids in the feature space.

During the training process, the driving scene parsing model is trained using both the source and target domain data alternately.
The source-domain prototype matrix $\boldsymbol{P}_s$ and target-domain prototype matrix $\boldsymbol{P}_t$ are computed based on the corresponding logits $\boldsymbol{L}_s$ and $\boldsymbol{L}_t$, and their respective ground-truth labels $\boldsymbol{Y}_s$ and pseudo labels $\boldsymbol{Y}_t$.
The final prototype matrix $\boldsymbol{P}_h \in \mathbb{R}^{D \times C} $ is updated by linearly combining $\boldsymbol{P}_s$ and $\boldsymbol{P}_t$ with a weight $\gamma$, as shown in the following equation:
\begin{equation}
\boldsymbol{P}_h \leftarrow \gamma \boldsymbol{P}_{s} + (1 - \gamma){\boldsymbol{P}_{t}},
\end{equation}
where $\gamma$ is a hyperparameter that controls the contribution of the source and target domain prototypes to the final prototype matrix.
In this study, $\gamma$ is set to 0.5 to ensure a balanced integration of prototypes from both domains, allowing the model to retain source-domain representations that capture inter-class differences while adapting to the feature characteristics of the target domain for improved cross-domain stability.

\subsubsection{Prototypical intra-class and inter-class interactions}
\label{sec.methodology.PtoP.intra}

To better characterize the feature distribution in the target domain and suppress the influence of noisy predictions, class prototypes are employed to represent the semantic structure of each category within the feature space. These prototypes serve as robust, noise-resistant summaries of class-specific features. By computing correlations between these prototypes, both inter-class separability and intra-class compactness are quantified, providing a structural perspective on feature distribution consistency.

\begin{figure*}[t!]
    \centering
    \includegraphics[width=0.99\textwidth]{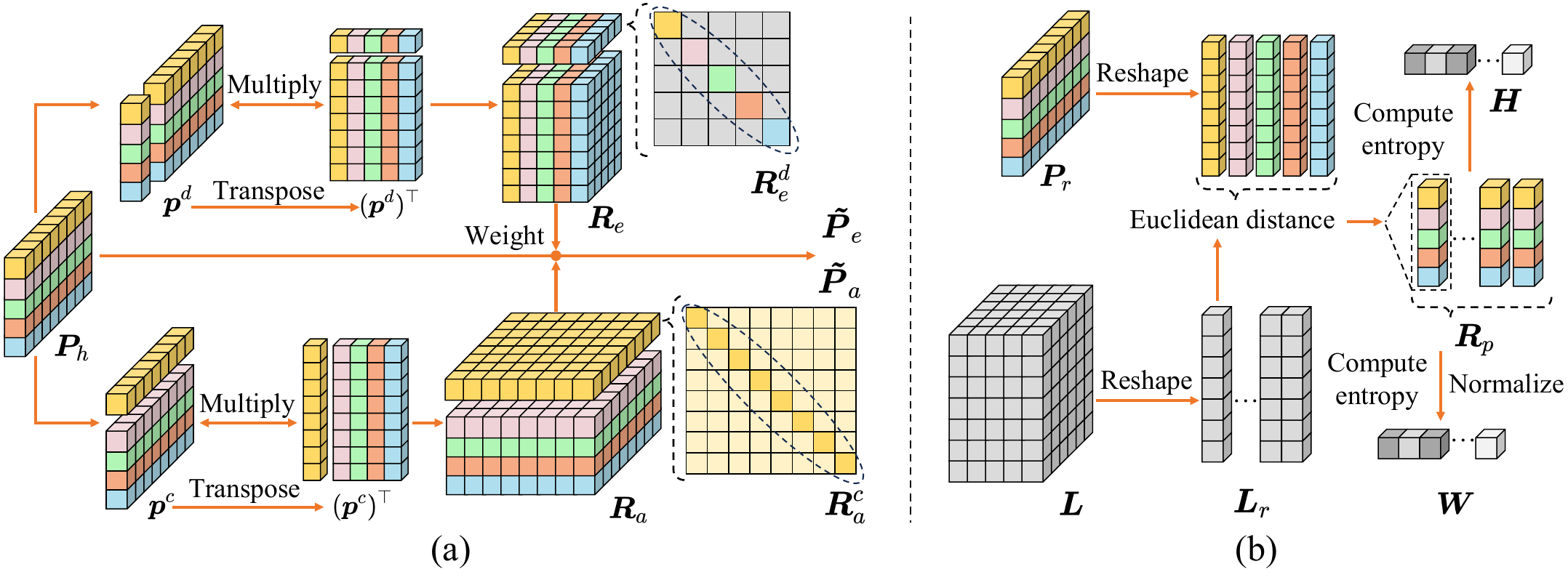}
    \caption{Illustration of prototype-based structural modeling and pixel-wise uncertainty estimation. (a) Generation of the inter-class and intra-class weighted prototypes, $\boldsymbol{\tilde P}_{e}$ and $\boldsymbol{\tilde P}_{a}$, via Prototype–Prototype interaction. (b) Estimation of the entropy map $\boldsymbol{H}$ and attention weight map $\boldsymbol{W}$ through Prototype–Pixel interaction.}
    \label{fig.matrix_interaction}
\end{figure*}

As illustrated in Fig.~\ref{fig.matrix_interaction}, the structural relationships between prototypes are first evaluated at the channel level. For the $d$-th feature channel, a prototype vector $\boldsymbol{p}^d \in \mathbb{R}^{1 \times C}$ is extracted from the prototype matrix $\boldsymbol{P}_h$. The corresponding inter-class interaction matrix $\boldsymbol{R}_{e}^{d} \in \mathbb{R}^{C \times C}$ is then computed as:
\begin{equation}
    \label{eq.sim_matri_inter}
    \boldsymbol{R}_{e}^{d} = \left({\boldsymbol{p}^d}\right)^\top \boldsymbol{p}^d,
\end{equation}
where each entry of $\boldsymbol{R}_{e}^{d}$ represents the correlation between the  $c$-th and $c'$-th classes in that specific channel. Specifically, the off-diagonal elements ($c \ne c'$) indicate cross-class similarity; high values indicate significant correlation between distinct semantic categories, reflecting potential semantic ambiguity. Conversely, the diagonal entries represent class-specific self-correlation, which should be emphasized to ensure class compactness. Finally, the matrices $\{\boldsymbol{R}_{e}^{d}\}_{d=1}^D$ from all channels are stacked to form the comprehensive inter-class interaction tensor $\boldsymbol{R}_{e} \in \mathbb{R}^{D \times C \times C}$.

To explicitly enforce this desirable property, the interaction matrices are normalized within each channel to enhance the prominence of diagonal values (self-correlation) relative to off-diagonal (cross-correlation) values. This normalization also ensures consistent scaling across channels, facilitating the formulation of regularization terms. For each channel, the normalized inter-class interaction $\boldsymbol{\tilde R}_{e}$ is defined as:
\begin{equation}
	\boldsymbol{\tilde R}_{e}^{(d,c,c')} ={\frac{{\boldsymbol{R}_{e}^{(d,c,c')}}}{{\boldsymbol{R}_{e}^{(d,c,c)}}+ {\epsilon}}},
\end{equation}
where $\epsilon$ is a small constant introduced to prevent division by zero.

Complementary to inter-class relationships, the proposed framework further models intra-class structural consistency to characterize the internal feature distribution of each semantic category across channels. Specifically, the $c$-th column of the prototype matrix $\boldsymbol{P}_h$ is extracted to form a prototype vector $\boldsymbol{p}^c \in \mathbb{R}^{1 \times D}$ representing the channel-wise feature response of the corresponding category. The intra-class interaction matrix is then defined as:
\begin{equation}
 \boldsymbol{R}_{a}^{c} = \left(\boldsymbol{p}^c\right)^\top \boldsymbol{p}^c,
\end{equation}
where $\boldsymbol{R}_{a}^{c}$ quantifies the correlation between the $d$-th and $d'$-th feature channels for the $c$-th class. Specifically, the diagonal elements ($d=d'$) represent self-correlations of individual channels, reflecting the strength of class-specific features, while the off-diagonal elements ($d \ne d'$) capture the correlations between different channels. A higher prominence of diagonal entries indicates that the feature representation of the class is compact and internally consistent.
Finally, the class-specific interaction matrices $\{\boldsymbol{R}_{a}^{c}\}_{c=1}^C$ are aggregated to construct a comprehensive intra-class interaction tensor $\boldsymbol{R}_{a} \in \mathbb{R}^{C \times D \times D}$.
To explicitly encourage internal consistency, the matrices are normalized across channels, yielding the normalized intra-class correlation $\boldsymbol{\tilde R}_{a}$:
\begin{equation}
	\boldsymbol{\tilde R}_{a}^{(c,d,d')} ={\frac{{\boldsymbol{R}_{a}^{(c,d,d')}}}{{\boldsymbol{R}_{a}^{(c,d,d)}}+ {\epsilon}}},
\end{equation}
where ${\epsilon}$ is a small constant introduced to prevent division by zero. 
This normalization emphasizes the relative importance of off-diagonal elements compared to dominant self-correlations and ensures a consistent scale for regularization across channels.

\subsubsection{Regularization based on prototypical interaction}
\label{sec.methodology.category-wise.constrain}

To effectively preserve the semantic structure embedded in the prototypes during cross-domain feature alignment, the normalized correlation matrices, $\boldsymbol{\tilde R}_{a}$ and $\boldsymbol{\tilde R}_{e}$, are leveraged as adaptive weights to recalibrate the original prototype matrix $\boldsymbol{P}_h$. 
The inter-class weighted prototype $\boldsymbol{\tilde P}_{e}\in \mathbb{R}^{D \times C}$ is obtained by integrating cross-class information, where each feature channel is scaled by its corresponding inter-class normalized correlation:
\begin{equation}
\boldsymbol{\tilde P}_{e}^{(d,c)} = \sum_{c'=1}^C \boldsymbol{\tilde R}_{e}^{(d,c,c')} \cdot \boldsymbol{P}_h^{(d,c')}.
\end{equation}
The intra-class weighted prototype $\boldsymbol{\tilde P}_{a}\in \mathbb{R}^{D \times C}$ is obtained by aggregating feature channels within each class, where each channel is weighted by its corresponding intra-class normalized correlation:
\begin{equation}
\boldsymbol{\tilde P}_{a}^{(d,c)} = \sum_{d'=1}^D \boldsymbol{\tilde R}_{a}^{(c,d,d')} \cdot \boldsymbol{P}_h^{(d',c)}.
\end{equation}
These weighted prototypes, $\boldsymbol{\tilde P}_{e}$ and $\boldsymbol{\tilde P}_{a}$, represent the inter-class ambiguity and intra-class inconsistency captured by the prototype correlations, respectively. Intuitively, $\boldsymbol{\tilde P}_{e}$ emphasizes features shared across classes that may cause semantic confusion, while $\boldsymbol{\tilde P}_{a}$ emphasizes channel-level interactions that reduce compactness within each class. 
To incorporate structural regularization during cross-domain feature alignment, the regularized prototype matrix is derived by refining the original $\boldsymbol{P}_h$ with weighted correlation terms as follows:
\begin{equation}
	\label{eq.reweigh_proto}
	\boldsymbol{P}_{r} = \boldsymbol{P}_h - \lambda_{e} {\boldsymbol{\tilde P}_{e}} - \lambda_{a} {\boldsymbol{\tilde P}_{a}},
\end{equation}
where the hyperparameters $\lambda_{e}$ and $\lambda_{a}$ control the strengths of the inter-class and intra-class regularization terms, respectively. 
Empirical evaluations show that setting both $\lambda_{e}$ and $\lambda_{a}$ to 0.1 achieves optimal performance. 
Under these settings, the structural corrections refine the prototype matrix $\boldsymbol{P}_h$, enhancing its semantic compactness within each class while preserving or improving its discriminability across classes. Serving as a robust anchor for contrastive learning, the refined $\boldsymbol{P}_{r}$ effectively integrates intrinsic structural properties into the alignment process, eliminating the need for additional supervision.

\subsection{Prototype-pixel interaction}
\label{sec.methodology.p2pixel.APT}

To enhance the domain adaptation process, a prototype-pixel interaction mechanism is introduced, which regularizes feature distributions by applying global structural constraints to local pixels, with the aim of reducing pixel-level semantic ambiguity.
As shown in Fig.~\ref{fig.matrix_interaction} (b), the logits $\boldsymbol{L}$ are reshaped into $\boldsymbol{L}_r \in \mathbb{R}^{ N \times C}$, where $N=H \times W$ denotes the total number of spatial positions. The semantic similarity between pixel-wise logits and class prototypes is quantified using the squared Euclidean distance:
\begin{equation}
	\boldsymbol{R}_{p}^{(n,c)} = \left\| \boldsymbol{L}_{r}^{(n,:)} - \boldsymbol{P}_{r}^{(:,c)} \right\|_2^2,
\end{equation}
where $\boldsymbol{R}_{p} \in \mathbb{R}^{N \times C}$ captures the similarity between each pixel-level logits and different semantic prototype vectors, serving as an indicator of classification confidence.

\subsubsection{Adaptive pseudo label threshold}
\label{sec.methodology.pixel_cluster.APT}

To identify reliable pixels for effective alignment, the prediction confidence is characterized by the entropy of the probability distribution derived from the distance matrix $\boldsymbol{R}_{p}$. The distance matrix is first normalized using the following softmax function:
\begin{equation}
	\boldsymbol{Q}^{(n,c)} = \frac{\exp\left(-\boldsymbol{R}_{p}^{(n,c)}\right)}{\sum_{c'=1}^{C} \exp\left(-\boldsymbol{R}_{p}^{(n,c')}\right)}.
\end{equation}
The pixel-wise prediction uncertainty is then computed as the entropy map $\boldsymbol{H}$ as follows:
\begin{equation}
    \boldsymbol{H}^{(n)} = -\sum_{c=1}^{C} \boldsymbol{Q}^{(n,c)} \log \boldsymbol{Q}^{(n,c)},
\end{equation}
where $\boldsymbol{H} \in \mathbb{R}^{N \times 1}$ quantifies the uncertainty of the pixel-level semantic predictions. Higher entropy values indicate greater uncertainty, reflecting more ambiguity in the pixel's predicted class, while lower entropy values suggest more confident and accurate predictions.
To eliminate unreliable samples, only the top-$\alpha$ fraction of pixels with the lowest entropy is retained to construct a binary mask for the subsequent feature distribution alignment process. 
This ratio-based selection strategy eliminates the requirement for manual thresholding, thereby enhancing the adaptability and flexibility of the method across diverse target domains with varying uncertainty distributions.

\subsubsection{Pixel-wise attention for contrastive learning}
\label{sec.methodology.pixel_cluster.attention_loss}

In addition to the selective filtering of unreliable samples, the entropy information is further leveraged as a soft attention mechanism to modulate the feature alignment process. By normalizing these entropy scores, the model assigns greater weight to pixels with higher ambiguity. This strategy prioritizes the refinement of challenging semantic regions, which typically necessitate more granular and precise domain alignment.
Specifically, the pixel-wise weights $\boldsymbol{W} \in \mathbb{R}^{N \times 1}$ are derived by normalizing the entropy map $\boldsymbol{H}$ relative to its global maximum value.
The normalized weight $\boldsymbol{W}$ serves as a relative measure of semantic uncertainty to guide the adaptation process.
This mechanism assigns higher weights to more ambiguous features, directing contrastive supervision to the most challenging regions, which are crucial for bridging the domain gap.

\subsubsection{Contrastive adaptation}

To facilitate semantically structured domain alignment, a prototype-based contrastive loss is developed to incorporate both structural regularization and pixel-wise attention. 
Specifically, the pixel-wise similarity between features and category-specific prototypes is computed for each spatial location $\boldsymbol{x}$ as a vector $\boldsymbol{p}_t=({p}_{t}^1,\cdots,{p}_{t}^C)^\top\in\mathbb{R}^{C}$, where each category-specific similarity score ${p}_{t}^c$ is calculated as follows:
\begin{equation}
    {p}_{t}^c = \frac{\exp\left( \boldsymbol{p}^{c}_{r} \cdot \boldsymbol{L}_t(\boldsymbol{x}) \cdot \boldsymbol{W}(\boldsymbol{x}) / \tau \right)}{\sum\nolimits_{k=1}^C \exp\left( \boldsymbol{p}_{r}^{k} \cdot \boldsymbol{L}_t(\boldsymbol{x}) \cdot \boldsymbol{W}(\boldsymbol{x}) / \tau \right)},
\end{equation}
where $\boldsymbol{x}$ denotes the spatial coordinate $(h, w)$ of a specific pixel and $\tau$ is the temperature coefficient that scales the similarity values. The vector $\boldsymbol{p}_t$ captures the pixel-wise similarity distribution across all $C$ categories.
Finally, the pixel-wise similarity vectors $\boldsymbol{p}_t$ across the entire image are combined into a similarity tensor 
$\boldsymbol{S}_t \in \mathbb{R}^{H \times W \times C}$.

To improve cross-domain feature alignment, higher weights are assigned to pixels near decision boundaries. This strategy amplifies the supervisory signal in ambiguous regions, ensuring that challenging spatial locations contribute more effectively to narrowing the domain gap.
Due to the unavailability of ground-truth annotations in the target domain, pseudo labels $\boldsymbol{Y}_t$ derived from model predictions are employed. To ensure their quality, an adaptive thresholding scheme is applied to filter out low-confidence predictions, thereby mitigating the risk of noisy supervision.
Leveraging the similarity scores and the filtered pseudo labels, the contrastive loss between the target-domain features and the class prototypes is defined as follows:
\begin{equation}
\mathcal{L}_{t} = - \mathcal{S}\left( \boldsymbol{H}_t \odot \log \ ({\boldsymbol{S}_t}) \right),
\end{equation}
where ${\boldsymbol{H}}_{t} \in \{0,1\}^{ H \times W \times C}$ denotes the one-hot encoding of pseudo labels $\boldsymbol{Y}_{t}$.
Additionally, intra-domain consistency within the source domain is enforced by defining a source-to-source contrastive loss:
\begin{equation}
\mathcal{L}_{s} = - \mathcal{S} \left(\boldsymbol{H}_s \odot \log ({\boldsymbol{S}_s}) \right),
\end{equation}
where ${\boldsymbol{H}}_{s} \in \{0,1\}^{ H \times W \times C}$ denotes the one-hot encoding of ground-truth labels $\boldsymbol{Y}_{s}$. The tensor ${\boldsymbol{S}_s}$ is computed in a similar manner to ${\boldsymbol{S}_t}$, capturing the pixel-wise similarity between source-domain features and class prototypes.
The overall contrastive adaptation objective integrates both directions:
\begin{equation}
\mathcal{L}_c = \mathcal{L}_{s} + \mathcal{L}_{t}.
\end{equation}
This formulation facilitates the alignment of pixel-wise features with their corresponding class prototypes while allocating greater importance to semantically uncertain regions. 
By integrating contrastive learning with an attention mechanism and adaptive pseudo labeling, the proposed method improves inter-class separability and enhances intra-class compactness, facilitating more robust cross-domain alignment.
To further improve adaptation performance, an additional self-training stage is incorporated. Following contrastive training, the model generates refined pseudo labels for target-domain images, which are subsequently employed to guide a self-supervised learning phase, thereby enhancing the model's performance in the target domain.

\begin{table*}[ht!]
	\centering
	\setlength{\heavyrulewidth}{2pt}
	\vspace{0.5em}
    \setlength{\tabcolsep}{0.55mm} 
    \tablesmallfont
	\caption{Quantitative comparison of state-of-the-art unsupervised domain adaptation methods for GTA5 $\to$ Cityscapes adaptation task based on per-class IoU ($\%$) and overall mIoU ($\%$).}
	\begin{tabular}{lc|ccccccccccccccccccc|cc}
		\toprule[0.9pt]
		Method &Publication & \rotatebox[origin=c]{90}{Road} & \rotatebox[origin=c]{90}{Sidewalk} & \rotatebox[origin=c]{90}{Building} & \rotatebox[origin=c]{90}{Wall} & \rotatebox[origin=c]{90}{Fence} & \rotatebox[origin=c]{90}{Pole} & \rotatebox[origin=c]{90}{Light} & \rotatebox[origin=c]{90}{Sign} & \rotatebox[origin=c]{90}{Vegetation} & \rotatebox[origin=c]{90}{Terrain} & \rotatebox[origin=c]{90}{Sky} & \rotatebox[origin=c]{90}{Person} & \rotatebox[origin=c]{90}{Rider} & \rotatebox[origin=c]{90}{Car} & \rotatebox[origin=c]{90}{Truck} & \rotatebox[origin=c]{90}{Bus} & \rotatebox[origin=c]{90}{Train} & \rotatebox[origin=c]{90}{Motorbike}& \rotatebox[origin=c]{90}{Bike} & \rotatebox[origin=c]{90}{mIoU} & \rotatebox[origin=c]{90}{Gain} \\
		\midrule
        Source Only~\cite{Chen2017a} &TPAMI'17 &53.8 &15.6 &69.3 &28.1 &18.8 &27.6 &34.9 &18.2 &82.5 &27.8 &71.6 &59.4 &35.3 &44.1 &25.9 &37.5 &0.1 &28.9 &24.9 &37.1 &+0.0\\
		PatchAlign~\cite{Tsai2019} &CVPR'19 &92.3&51.9 &82.1 &29.2 &25.1 &24.5 &33.8 &33.0 &82.4 &32.8 &82.2 &58.6 &27.2 &84.3 &33.4 &46.3 &2.2 &29.5 &32.3 &46.5 &+9.4\\
        ADVENT~\cite{Vu2019} &CVPR'19&89.4 &33.1 &81.0 &26.6 &26.8 &27.2 &33.5 &24.7 &83.9 &36.7 &78.8 &58.7 &30.5 &84.8 &38.5 &44.5 &1.7 &31.6 &32.4 &45.5 &+8.4\\
        AdaptSeg~\cite{Tsai2018} &CVPR'18&86.5 &36.0 &79.9 &23.4 &23.3 &23.9 &35.2 &14.8 &83.4 &33.3 &75.6 &58.5 &27.6 &73.7 &32.5 &35.4 &3.9 &30.1 &28.1 &42.4 &+5.3\\
        BDL~\cite{Li2019} &CVPR'19&91.0 &44.7 &84.2 &34.6 &27.6 &30.2 &36.0 &36.0 &85.0 &43.6 &83.0 &58.6 &31.6 &83.3 &35.3 &49.7 &3.3 &28.8 &35.6 &48.5&+11.4 \\
        UIDA~\cite{Pan2020} &CVPR'20&90.6 &37.1 &82.6 &30.1 &19.1 &29.5 &32.4 &20.6 &85.7 &40.5 &79.7 &58.7 &31.1 &86.3 &31.5 &48.3 &0.0 &30.2 &35.8 &46.3&+9.2 \\
        LTIR~\cite{Kim2020} &CVPR'20&92.9 &55.0 &85.3 &34.2 &31.1 &34.9 &40.7 &34.0 &85.2 &40.1 &87.1 &61.0 &31.1 &82.5 &32.3 &42.9 &0.3 &36.4 &46.1 &50.2&+13.1 \\
        PIT~\cite{Lv2020} &CVPR'20&87.5 &43.4 &78.8 &31.2 &30.2 &36.3 &39.9 &42.0 &79.2 &37.1 &79.3 &65.4 &37.5 &83.2 &{46.0} &45.6 &25.7 &23.5 &49.9 &50.6&+13.5 \\
        LSE~\cite{Subhani2020} &ECCV'20&90.2 &40.0 &83.5 &31.9 &26.4 &32.6 &38.7 &37.5 &81.0 &34.2 &84.6 &61.6 &33.4 &82.5 &32.8 &45.9 &6.7 &29.1 &30.6 &47.5&+10.4 \\
        WeakSeg~\cite{Paul2020} &ECCV'20&91.6 &47.4 &84.0 &30.4 &28.3 &31.4 &37.4 &35.4 &83.9 &38.3 &83.9 &61.2 &28.2 &83.7 &28.8 &41.3 &8.8 &24.7 &46.4 &48.2&+11.1 \\
        CrCDA~\cite{Huang2020} &ECCV'20&92.4 &55.3 &82.3 &31.2 &29.1 &32.5 &33.2 &35.6 &83.5 &34.8 &84.2 &58.9 &32.2 &84.7 &40.6 &46.1 &2.1 &31.1 &32.7 &48.6 &+11.5 \\
        FADA~\cite{Wang2020a} &ECCV'20&92.5 &47.5 &85.1 &37.6 &32.8 &33.4 &33.8 &18.4 &85.3 &37.7 &83.5 &63.2 &39.7 &87.5 &32.9 &47.8 &1.6 &34.9 &39.5 &49.2&+12.1 \\
        IAST~\cite{Mei2020} &ECCV'20&94.1 &58.8 &85.4 &39.7 &29.2 &25.1 &43.1 &34.2 &84.8 &34.6 &88.7 &62.7 &30.3 &{87.6} &42.3 &50.3 &24.7 &35.2 &40.2 &52.2&+15.1 \\
        ASA~\cite{Zhou2020} &TIP'21&89.2 &27.8 &81.3 &25.3 &22.7 &28.7 &36.5 &19.6 &83.8 &31.4 &77.1 &59.2 &29.8 &84.3 &33.2 &45.6 &16.9 &34.5 &30.8 &45.1&+8.0 \\
        CLAN~\cite{Luo2021} &TPAMI'21&88.7 &35.5 &80.3 &27.5 &25.0 &29.3 &36.4 &28.1 &84.5 &37.0 &76.6 &58.4 &29.7 &81.2 &38.8 &40.9 &5.6 &32.9 &28.8 &45.5&+8.4 \\
        DACS~\cite{Tranheden2021} &WACV'21&89.9 &39.7 &87.9 &39.7 &39.5 &38.5 &46.4 &{52.8} &88.0 &44.0 &88.8 &{67.2} &35.8 &84.5 &45.7 &50.2 &0.0 &27.3 &34.0 &52.1&+15.0 \\
        RPLL~\cite{Zheng2021} &IJCV'21&90.4 &31.2 &85.1 &36.9 &25.6 &37.5 &{48.8} &48.5 &85.3 &34.8 &81.1 &64.4 &36.8 &86.3 &34.9 &52.2 &1.7 &29.0 &44.6 &50.3&+13.2 \\
        DAST~\cite{Yu2021} &AAAI'21&92.2 &49.0 &84.3 &36.5 &28.9 &33.9 &38.8 &28.4 &84.9 &41.6 &83.2 &60.0 &28.7 &87.2 &45.0 &45.3 &7.4 &33.8 &32.8 &49.6&+12.5 \\
        ConTrans~\cite{Lee2021} &AAAI'21 &{95.3} &{65.1} &84.6 &33.2 &23.7 &32.8 &32.7 &36.9 &86.0 &41.0 &85.6 &56.1 &25.9 &86.3 &34.5 &39.1 &11.5 &28.3 &43.0 &49.6&+12.5 \\
        CIRN~\cite{Gao2021} &AAAI'21&91.5 &48.7 &85.2 &33.1 &26.0 &32.3 &33.8 &34.6 &85.1 &43.6 &86.9 &62.2 &28.5 &84.6 &37.9 &47.6 &0.0 &35.0 &36.0 &49.1&+12.0 \\
        CLST~\cite{Marsden2022} &IJCNN'21&92.8 &53.5 &86.1 &39.1 &28.1 &28.9 &43.6 &39.4 &84.6 &35.7 &88.1 &63.9 &38.3 &86.0 &41.6 &50.6 &0.1 &30.4 &51.7 &51.6&+14.5 \\
        MetaCorrect~\cite{Guo2021} &CVPR'21&92.8 &58.1 &86.2 &39.7 &33.1 &36.3 &42.0 &38.6 &85.5 &37.8 &87.6 &62.8 &31.7 &84.8 &35.7 &50.3 &2.0 &36.8 &48.0 &52.1&+15.0 \\
        ProDA~\cite{Zhang2021} &CVPR'21&91.5 &52.3 &82.9 &42.0 &{35.7} &40.0 &44.4 &43.2 &87.0 &43.8 &79.5 &66.4 &31.3 &86.7 &41.1 &52.5 &0.0 &{45.4} &53.8 &53.7 &+16.6 \\
        UPLR~\cite{Wang2021} &ICCV'21&90.5 &38.7 &86.5 &41.1 &32.9 &{40.5} &48.2 &42.1 &86.5 &36.8 &84.2 &64.5 &38.1 &87.2 &34.8 &50.4 &0.2 &41.8 &{54.6} &52.6&+15.5 \\
        EHTDI~\cite{EHTDI_2022} &ACM MM'22 &95.4 &68.8 &88.1 &37.1 &\textbf{41.4} &42.5 &45.7 &60.4 &87.3 &42.6 &86.8 &67.4 &38.6 &90.5 &\textbf{66.7} &61.4 &0.3 &39.4 &56.1 &58.8 &+21.7\\
        EIC~\cite{chung2023exploiting} &WACV'23 &90.8 &47.2 &86.8 &41.5 &29.4 &35.7 &42.4 &37.4 &86.0 &42.1 &88.3 &63.7 &35.6 &85.1 &43.8 &54.6 &0.0 &33.6 &47.8 &52.2 &+15.1\\
        ARAS~\cite{cao2023adaptive} &TCSVT'23 &91.9 &45.2 &81.8 &21.9 &25.6 &35.5 &41.5 &33.4 &85.1 &34.8 &73.8 &62.5 &31.6 &85.9 &33.8 &42.5 &7.3 &33.8 &42.8 &47.9 &+10.8 \\
        ADPL-Dual~\cite{ADPL-Dual_2023} &TPAMI'23 &93.4 &60.6 &87.5 &45.3 &32.6 &37.3 &43.3 &55.5 &87.2 &44.8 &88.0 &64.5 &34.2 &88.3 &52.6 &61.8 &49.8 &41.8 &59.4 &59.4 &+22.3 \\
        SePiCo~\cite{SePiCo_2023} &TPAMI'23 &95.2 &67.8 &88.7 &41.4 &38.4 &43.4 &\textbf{55.5} &\textbf{63.2} &88.6 &46.4 &88.3 &\textbf{73.1} &\textbf{49.0} &\textbf{91.4} &{63.2} &60.4 &0.0 &45.2 &\textbf{60.0} &61.0 &+23.9 \\
        FREDOM~\cite{FREDOM_2023} &CVPR'23 &90.9 &54.1 &87.8 &44.1 &32.6 &\textbf{45.2} &51.4 &57.1 &88.6 &42.6 &89.5 &68.8 &40.0 &89.7 &58.4 &62.6 &55.3 &47.7 &58.1 &61.3 &+24.2\\
        PBAL~\cite{ren2023prototypical} &TMM'24 &88.3 &38.2 &85.3 &32.5 &29.0 &37.5 &43.7 &40.5 &86.7 &33.5 &84.8 &65.9 &28.6 &85.5 &33.3 &31.3 &{26.8} &20.4 &48.6 &49.5 & +12.4\\
        CACP~\cite{CACP_2025} &TMM'25 &\textbf{95.5} &\textbf{70.4} &88.1 &\textbf{52.3} &36.2 &44.4 &53.1 &57.2 &89.6 &51.3 &\textbf{90.0} &71.3 &42.9 &88.5 &32.7 &61.3 &\textbf{65.0} &\textbf{53.1} &47.0 &\textbf{62.6} &\textbf{+25.5} \\
		\midrule
		\textbf{SPR (Ours)} &SCIS'26 &90.7 &43.2 &91.0 &40.3 &29.6 &32.9 &41.5 &38.0 &91.0 &50.9 &{89.0} &63.1 &{38.2} &81.7 &43.3 &60.2 &10.8 &31.2 &37.5 &53.1 &+16.0 \\
        \textbf{SPR+ST (Ours)} &SCIS'26 &90.4 &40.7 &\textbf{91.5} &{47.1} &30.8 &32.8 &41.0 &33.2 &\textbf{91.1} &\textbf{58.9} &87.7 &61.3 &35.0 &79.1 &51.5 &\textbf{65.6} &0.2 &43.7 &52.6 &{54.4} &{+17.3}\\
        \bottomrule[0.9pt]
	\end{tabular}
	\label{tab.gtav2city}
\end{table*}

\section{Experiments}
\label{sec.exp}

\subsection{Datasets and evaluation metrics}
\label{sec.exp.dataset}

Following previous works~\cite{Jiang2022,Wang2020a,ren2023prototypical,cao2023adaptive,chung2023exploiting}, the proposed method is evaluated on three widely used unsupervised domain adaptation benchmarks for driving scene parsing: (1) GTA5 $\to$ Cityscapes, (2) SYNTHIA $\to$ Cityscapes, and (3) Cityscapes $\to$ ACDC.
The GTA5 dataset~\cite{Richter2016} includes 24,966 synthetic images with a resolution of 1,914$\times$1,052 pixels and contains 19 semantic classes aligned with the Cityscapes dataset. The SYNTHIA dataset~\cite{Ros2016} contains 9,400 synthetic images with a resolution of 1,280$\times$760 pixels. The Cityscapes dataset~\cite{Cordts2016} contains 2,975 real-world images for training, 500 real-world images for validation, and 1,525 real-world images for testing, all with a resolution of 2,048$\times$1,024 pixels. 
The Adverse Conditions Dataset with Correspondences (ACDC)~\cite{acdc_data} is specifically designed to capture adverse visual conditions in the real world, including fog, night, rain, and snow, and shares the same semantic classes as the Cityscapes dataset.
In this study, mean intersection over union (mIoU) is used as the primary evaluation metric, providing a comprehensive quantification of overall scene parsing performance across semantic classes. 

\subsection{Implementation details}
\label{sec.exp.implementation_details}

Consistent with previous studies~\cite{Jiang2022,Wang2020a,ren2023prototypical,cao2023adaptive,chung2023exploiting}, the DeepLab-v2 model~\cite{Chen2017a} with a ResNet-101~\cite{He2016} encoder is adopted as the baseline segmentation model. 
All methods involved in the comparison are initialized with weights pre-trained on the ImageNet dataset~\cite{Deng2009}.
Atrous spatial pyramid pooling (ASPP)~\cite{Chen2017a} is used after the final encoder layer with dilation rates of \{6, 12, 18, 24\}. An up-sampling layer is then employed to generate the segmentation outputs of the driving scene parsing model. 
To simulate real-world deployment scenarios, the source-domain dataset is partitioned into training, validation, and test sets in a ratio of 6:1:3, and the initial weights are selected based on the validation set. 
All methods involved in the comparison are implemented using PyTorch and executed on a heterogeneous system equipped with an Intel(R) Xeon(R) Gold 6154 CPU and a single NVIDIA RTX 3090 GPU with 24 GB memory. 
The proposed SPR model is optimized using the stochastic gradient descent optimizer with an initial learning rate of $2.5\times10^{-4}$, a momentum of 0.9, and a weight decay of $5.0\times10^{-4}$. The learning rate is adjusted according to a polynomial decay schedule with a power of 0.9.

\subsection{Comparisons with state-of-the-art unsupervised domain adaptation methods}
\label{sec.exp.State-of-the-Arts}

\begin{figure*}[t!]
    \centering
    \includegraphics[width=0.99\textwidth]{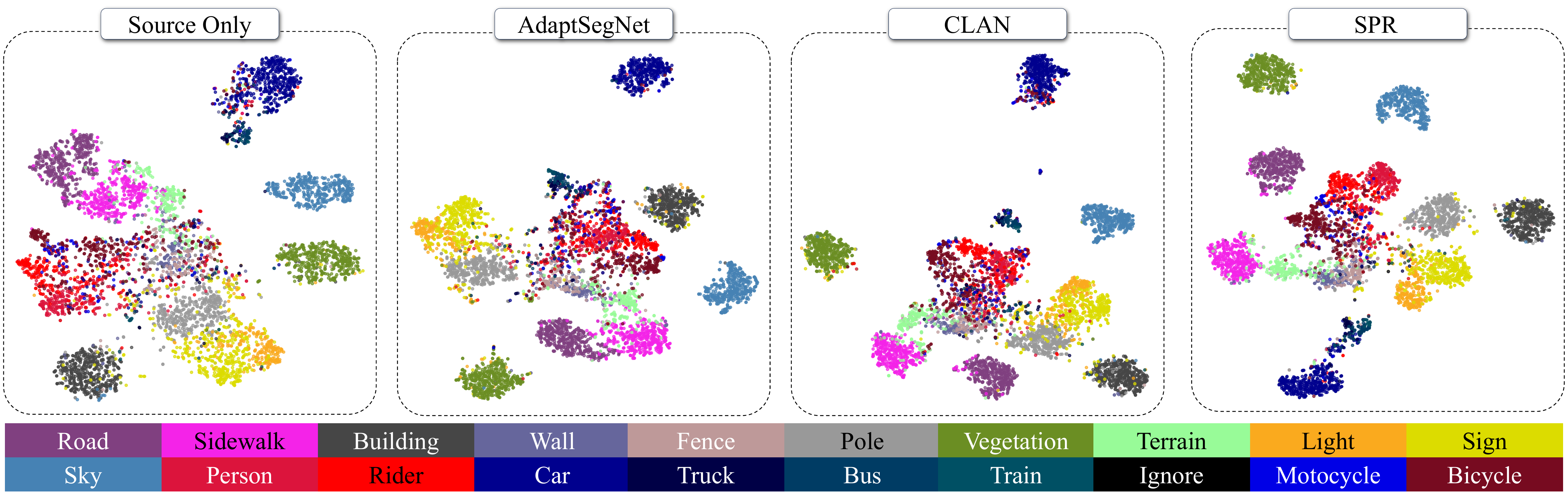}
    \caption{Feature cluster visualization in the output space using t-SNE~\cite{VanderMaaten2008}. }
    \label{fig.tsne}
\end{figure*}

\subsubsection{Experimental results on GTA5 → Cityscapes adaptation task}
\label{sec.exp.gtavtocityscapes}

\begin{figure*}[t!]
    \centering
    \includegraphics[width=0.99\textwidth]{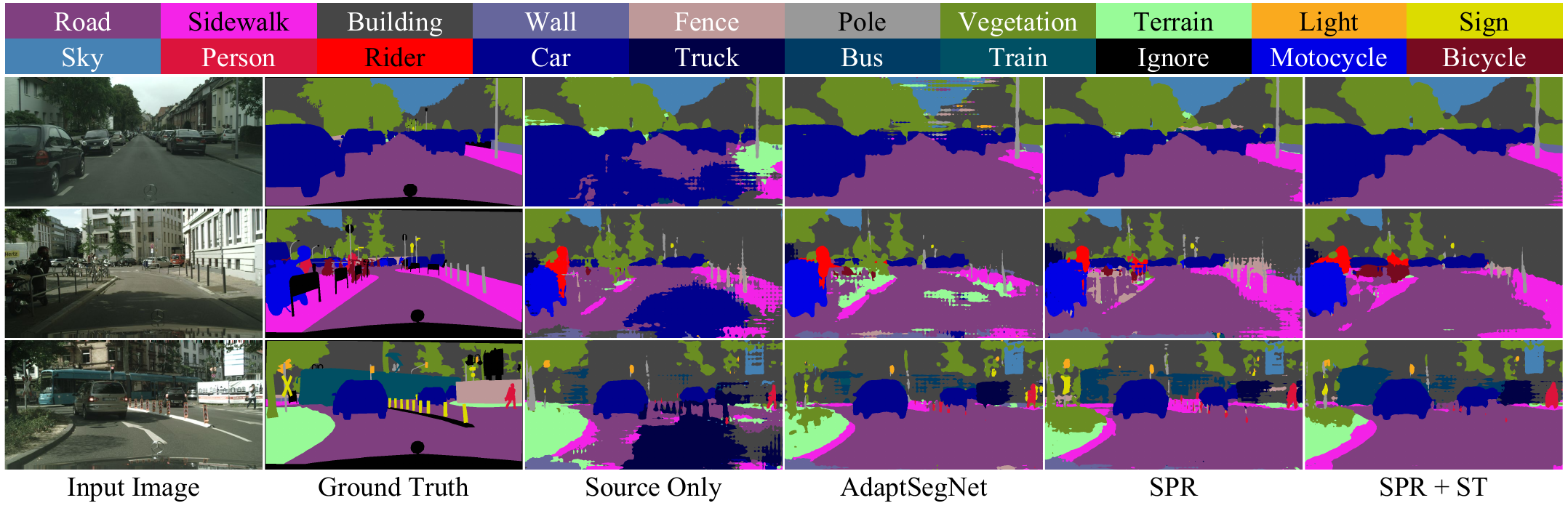}
    \caption{Qualitative results obtained on the GTA5 $\to$ Cityscapes adaptation task. From left to right, the figure displays the input target-domain image and its corresponding ground truth, followed by the segmentation results predicted by four different methods: the source-only baseline model, AdaptSegNet~\cite{Tsai2018}, SPR, and SPR with the self-training strategy.}
    \label{fig.gtav_pic}
\end{figure*}

The quantitative and qualitative experimental results are presented in Table~\ref{tab.gtav2city} and Fig.~\ref{fig.gtav_pic}. These results suggest that the proposed SPR method outperforms the baseline and various existing methods, particularly those based on adversarial learning~\cite{Wang2020a,Luo2021,Yu2021,Li2019}. 
It should be noted here that the proposed method acts primarily on the feature adaptation stage and is therefore independent of, yet complementary to, improvements obtained through self-training (ST) strategies.
It can be effectively integrated with self-training methods to further enhance performance. 
Specifically, SPR achieves an mIoU of 53.1$\%$, representing a substantial 16.0$\%$ improvement over the baseline. When integrated with self-training, SPR further boosts performance to 54.4$\%$, supporting our hypothesis that explicitly modeling inter-class separability and intra-class compactness is essential for feature alignment.
Furthermore, while several state-of-the-art (SoTA) methods optimize for long-tail categories, SPR demonstrates competitive performance on large-scale categories. In particular, SPR+ST outperforms recent methods such as CACP~\cite{CACP_2025} and SePiCo~\cite{SePiCo_2023} by achieving 91.5$\%$ IoU on building, 91.1$\%$ on vegetation, 58.9$\%$ on terrain, and 65.6$\%$ on bus. These results indicate that SPR extends traditional prototype-based methods by introducing structural regularization, yielding superior semantic consistency for major scene components.

To further evaluate SPR, qualitative segmentation results are presented in Fig.~\ref{fig.gtav_pic}. Notably, the driving scene parsing results exhibit significant improvement over the source-only baseline. Additionally, feature distributions in the output space are visualized using t-SNE~\cite{VanderMaaten2008}, as illustrated in Fig.~\ref{fig.tsne}. These visualizations demonstrate that SPR induces more compact intra-class clusters and clearer inter-class separation, thereby validating the effectiveness of the structural regularization strategy.

\begin{table*}[ht!]
	\centering
	\setlength{\heavyrulewidth}{2pt}
	\vspace{0.5em}
    \setlength{\tabcolsep}{0.6mm} 
    \tablesmallfont
	\caption{Quantitative comparison of state-of-the-art unsupervised domain adaptation methods for SYNTHIA $\to$ Cityscapes adaptation task based on per-class IoU ($\%$) and overall mIoU ($\%$). ${\ast}$ indicates challenging categories; mIoU$^{\ast}$ ($\%$) is calculated over the remaining categories.}
	\begin{tabular}{lc|cccccccccccccccc|cc|cc}
		\toprule[0.9pt]
		Method &Publication
        & \rotatebox[origin=c]{90}{Road} & \rotatebox[origin=c]{90}{Sidewalk} & \rotatebox[origin=c]{90}{Building} 
        & \rotatebox[origin=c]{90}{{Wall}$^{\ast}$} & \rotatebox[origin=c]{90}{Fence$^{\ast}$} & \rotatebox[origin=c]{90}{Pole$^{\ast}$} 
        & \rotatebox[origin=c]{90}{Light} & \rotatebox[origin=c]{90}{Sign} & \rotatebox[origin=c]{90}{Vegetation} 
        & \rotatebox[origin=c]{90}{Sky} & \rotatebox[origin=c]{90}{Person} & \rotatebox[origin=c]{90}{Rider} 
        & \rotatebox[origin=c]{90}{Car} & \rotatebox[origin=c]{90}{Bus} & \rotatebox[origin=c]{90}{Motorbike}
        & \rotatebox[origin=c]{90}{Bike} & \rotatebox[origin=c]{90}{mIoU} & \rotatebox[origin=c]{90}{Gain} & \rotatebox[origin=c]{90}{mIoU$^{\ast}$} & \rotatebox[origin=c]{90}{Gain}\\
		\midrule
        Source Only~\cite{Chen2017a} &TPAMI'17 &55.6 &23.8 &74.6 &9.2 &0.2 &24.4 &6.1 &12.1 &74.8 &79.0 &55.3 &19.1 &39.6 &23.3 &13.7 &25.0 &33.5 &0.0 &38.6 &+0.0\\
        PatchAlign~\cite{Tsai2019} &CVPR'19 &82.4 &38.0 &78.6 &- &- &- &9.9 &10.5 &78.2 &80.5 &53.5 &19.6 &67.0 &29.5 &21.6 &31.3 &- &- &46.2 &+7.6\\
        ADVENT~\cite{Vu2019} &CVPR'19 &85.6 &42.2 &79.7 &8.7 &0.4 &25.9 &5.4 &8.1 &80.4 &84.1 &57.9 &23.8 &73.3 &36.4 &14.2 &33.0 &41.2 &+7.7 &48.0 &+9.4\\
        AdaptSeg~\cite{Tsai2018} &CVPR'18 &84.3 &42.7 &77.5 &- &-&- &4.7 &7.0 &77.9 &82.5 &54.3 &21.0 &72.3 &32.2 &18.9 &32.3 &-&- &46.7 &+8.1\\
        BDL~\cite{Li2019} &CVPR'19 &86.0 &46.7 &80.3 &- &- &- &14.1 &11.6 &79.2 &81.3 &54.1 &27.9 &73.7 &42.2 &25.7 &45.3 &- &-&51.4 &+12.8\\
        UIDA~\cite{Pan2020} &CVPR'20 &84.3 &37.7 &79.5 &5.3 &0.4 &24.9 &9.2 &8.4 &80.0 &84.1 &57.2 &23.0 &78.0 &38.1 &20.3 &36.5 &41.7 &+8.2 &48.9 &+10.3\\
        LTIR~\cite{Kim2020}&CVPR'20 &92.6 &53.2 &79.2 &- &- &- &1.6 &7.5 &78.6 &84.4 &52.6 &20.0 &82.1 &34.8 &14.6 &39.4 &- &- &49.3 &+10.7\\ 
        PIT~\cite{Lv2020} &CVPR'20 &83.1 &27.6 &81.5 &8.9 &0.3 &21.8 &26.4 &33.8 &76.4 &78.8 &64.2 &27.6 &79.6 &31.2 &31.0 &31.3 &44.0 &+10.5 &51.8 &+13.2\\
        LSE~\cite{Subhani2020} &ECCV'20 &82.9 &43.1 &78.1 &9.3 &0.6 &28.2 &9.1 &14.4 &77.0 &83.5 &58.1 &25.9 &71.9 &38.0 &29.4 &31.2 &42.5 &+9.0 &49.4 &+10.8\\
        WeakSeg~\cite{Paul2020} &ECCV'20 &92.0 &53.5 &80.9 &11.4 &0.4 &21.8 &3.8 &6.0 &81.6 &84.4 &60.8 &24.4 &80.5 &39.0 &26.0 &41.7 &44.3 &+10.8 &51.9 &+13.3\\
        CrCDA~\cite{Huang2020} &ECCV'20 &86.2 &44.9 &79.5 &8.3 &0.7 &27.8 &9.4 &11.8 &78.6 &86.5 &57.2 &26.1 &76.8 &39.9 &21.5 &32.1 &42.9 &+9.4 &50.0 &+11.4\\
        FADA~\cite{Wang2020a} &ECCV'20 &84.5 &40.1 &83.1 &4.8 &0.0 &34.3 &20.1 &27.2 &84.8 &84.0 &53.5 &22.6 &85.4 &43.7 &26.8 &27.8 &45.2 &+11.7 &52.5 &+13.9\\
        IAST~\cite{Mei2020} &ECCV'20 &81.9 &41.5 &83.3 &17.7 &4.6 &32.3 &30.9 &28.8 &83.4 &85.0 &65.5 &30.8 &86.5 &38.2 &33.1 &52.7 &49.8 &+16.3 &57.0 &+18.4\\
        ASA~\cite{Zhou2020} &TIP'21 &91.2 &48.5 &80.4 &3.7 &0.3 &21.7 &5.5 &5.2 &79.5 &83.6 &56.4 &21.9 &80.3 &36.2 &20.0 &32.9 &41.7 &+8.2 &49.3 &+10.7\\
        CLAN~\cite{Luo2021} &TPAMI'21 &82.7 &37.2 &81.5 &- &- &- &17.1 &13.1 &81.2 &83.3 &55.5 &22.1 &76.6 &30.1 &23.5 &30.7 &- &- &48.8 &+10.2\\
        DACS~\cite{Tranheden2021} &WACV'21 &80.6 &25.1 &81.9 &21.5 &2.9 &37.2 &22.7 &24.0 &83.7 &\textbf{90.8} &67.6 &38.3 &82.9 &38.9 &28.5 &47.6 &48.3 &+14.8 &54.8 &+16.2\\
        RPLL~\cite{Zheng2021} &IJCV'21 &87.6 &41.9 &83.1 &14.7 &1.7 &36.2 &31.3 &19.9 &81.6 &80.6 &63.0 &21.8 &86.2 &40.7 &23.6 &53.1 &47.9 &+14.4 &54.9 &+16.3\\
        DAST~\cite{Yu2021} &AAAI'21 &87.1 &44.5 &82.3 &10.7 &0.8 &29.9 &13.9 &13.1 &81.6 &86.0 &60.3 &25.1 &83.1 &40.1 &24.4 &40.5 &45.2 &+11.7 &52.5 &+13.9\\
        ConTrans~\cite{Lee2021} &AAAI'21 &93.3 &{54.0} &81.3 &14.3 &0.7 &28.8 &21.3 &22.8 &82.6 &83.3 &57.7 &22.8 &83.4 &30.7 &20.2 &47.2 &46.5 &+13.0 &53.9 &+15.3\\
        CIRN~\cite{Gao2021} &AAAI'21 &85.8 &40.4 &80.4 &4.7 &1.8 &30.8 &16.4 &18.6 &80.7 &80.4 &55.2 &26.3 &83.9 &43.8 &18.6 &34.3 &43.9 &+10.4 &51.1 &+12.5\\
        CLST~\cite{Marsden2022} &IJCNN'21 &88.0 &49.2 &82.2 &16.3 &0.4 &29.2 &31.8 &23.9 &84.1 &88.0 &59.1 &27.2 &85.5 &46.4 &28.9 &56.5 &49.8 &+16.3 &57.8 &+19.2\\ 
        MetaCorrect~\cite{Guo2021} &CVPR'21 &92.6 &52.7 &81.3 &8.9 &2.4 &28.1 &13.0 &7.3 &83.5 &85.0 &60.1 &19.7 &84.8 &37.2 &21.5 &43.9 &45.1 &+11.6 &52.5 &+13.9\\
        ProDA~\cite{Zhang2021} &CVPR'21 &87.1 &44.0 &83.2 &26.9 &0.0 &42.0 &45.8 &34.2 &86.7 &81.3 &68.4 &22.1 &\textbf{87.7} &{50.0} &31.4 &38.6 &{51.9} &+18.4 &58.5 &+19.9\\
        UPLR~\cite{Wang2021} &ICCV'21 &79.4 &34.6 &83.5 &19.3 &2.8 &35.3 &32.1 &26.9 &78.8 &79.6 &66.6 &30.3 &86.1 &36.6 &19.5 &56.9 &48.0 &+14.5 &54.6 &+16.0\\
        EHTDI~\cite{EHTDI_2022} &ACM MM'22 &93.0 &\textbf{69.8} &84.0 &{36.6} &\textbf{9.1} &39.7 &42.2 &43.8 &\textbf{88.2} &88.1 &68.3 &29.0 &85.5 &54.1 &37.1 &56.3 &57.8 &+24.3 &64.6 &+26.0\\
        ADPL-Dual~\cite{ADPL-Dual_2023} &TPAMI'23 &86.1 &38.6 &85.9 &29.7 &1.3 &36.6 &41.3 &47.2 &85.0 &90.4 &67.5 &44.3 &87.4 &57.1 &43.9 &51.4 &55.9 &+22.4 &63.6 &+25.0\\
        EIC~\cite{chung2023exploiting} &WACV'23 &77.5 &32.3 &82.6 &25.5 &1.9 &34.6 &33.6 &32.4 &81.7 &85.1 &63.8 &31.8 &82.3 &35.2 &31.9 &54.6 &49.2 &+15.7 &55.7 &+17.1\\
        ARAS~\cite{cao2023adaptive} &TCSVT'23 &85.6 &39.2 &79.9 &15.5 &0.3 &32.2 &19.3 &23.9 &79.1 &81.7 &61.1 &19.3 &82.9 &25.7 &10.6 &51.9 &44.3 &+10.8 &50.8 &+12.2\\
        SePiCo~\cite{SePiCo_2023} &TPAMI'23 &77.0 &35.3 &85.1 &23.9 &3.4 &38.0 &51.0 &\textbf{55.1} &85.6 &80.5 &73.5 &\textbf{46.3} &87.6 &69.7 &50.9 &\textbf{66.5} &58.1 &+24.6 &66.5 &+27.9\\
        FREDOM~\cite{FREDOM_2023} &CVPR'23 &86.0 &46.3 &\textbf{87.0} &33.3 &5.3 &\textbf{48.7} &\textbf{53.4} &46.8 &87.1 &89.1 &71.2 &38.1 &87.1 &54.6	&51.3 &59.9 &59.1 &+25.6 &- &- \\
        PBAL~\cite{ren2023prototypical} &TMM'24 &85.3 &42.5 &81.7 &10.2 &0.1 &36.8 &23.6 &31.8 &85.1 &87.6 &64.1 &27.7 &85.2 &31.4 &23.0 &35.6 &47.0 &+13.5 &54.2 &+15.6\\
        CACP~\cite{CACP_2025} &TMM'25 &77.8 &37.5 &85.9 &\textbf{37.4} &0.0 &44.4 &52.7 &50.9 &\textbf{88.2} &88.0 &\textbf{75.1} &38.2 &87.4 &\textbf{78.9} &\textbf{56.0} &55.7 &\textbf{59.6} &\textbf{+26.1} &\textbf{67.1} &\textbf{+28.5}\\ 
               
        \midrule

		\textbf{SPR (Ours)} &SCIS'26 &96.2 &48.0 &86.0 &17.6 &1.1 &29.5 &17.0 &14.7 &81.4 &81.0 &60.3 &25.9 &85.1 &38.9 &27.5 &51.4 &47.6 &+14.1 &54.9 &+16.3 \\
        \textbf{SPR+ST (Ours)} &SCIS'26 &\textbf{97.9} &52.6 &{86.9} &15.7 &2.1 &30.0 &25.3 &15.8 &83.6 &83.1 &63.7 &31.2 &86.7 &49.5 &{35.3} &{60.6} &51.2 &+17.7 &{59.4} &{+20.8} \\
        \bottomrule[0.9pt]
	\end{tabular}
	\label{tab.synthia}
\end{table*}

\begin{figure*}[t!]
    \centering
    \includegraphics[width=0.99\textwidth]{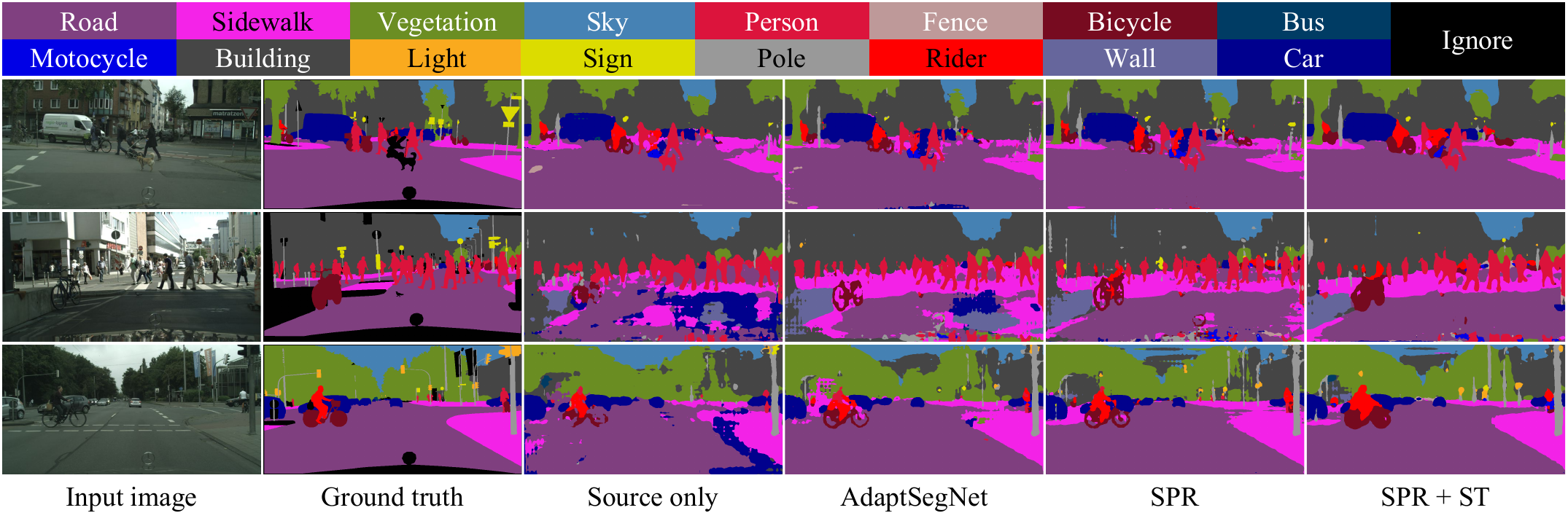}
    \caption{Qualitative results on the SYNTHIA $\to$ Cityscapes adaptation task. From left to right, the figure displays the input target-domain image and its corresponding ground truth, followed by the segmentation results predicted by four different methods: the source-only baseline model, AdaptSegNet~\cite{Tsai2018}, SPR, and SPR with the self-training strategy.}
    \label{fig.synthia_pic}
\end{figure*}

\subsubsection{Experimental results on SYNTHIA → Cityscapes adaptation task}
\label{sec.exp.synthiatocityscapes}

The quantitative and qualitative experimental results are presented in Table~\ref{tab.synthia} and Fig.~\ref{fig.synthia_pic}.
The domain discrepancy in the SYNTHIA $\to$ Cityscapes adaptation task is significantly more pronounced than in the GTA5 $\to$ Cityscapes adaptation task due to substantial differences in spatial layout.  As shown in Table~\ref{tab.synthia}, SPR effectively addresses these alignment difficulties, delivering substantial improvements over the baseline model. While traditional methods~\cite{Yu2021,Luo2021} based on adversarial learning often struggle with this heightened discrepancy and yield limited gains, the proposed SPR shows superior adaptability. Notably, when integrated with self-training strategies, SPR+ST achieves a remarkable performance boost, reaching an mIoU$^{\ast}$ of 59.4$\%$. This confirms that the feature alignment strategy provides a solid foundation for subsequent refinement stages.
Compared to existing UDA methods, SPR offers enhanced consistency for major, safety-critical categories. While SoTA methods such as CACP~\cite{CACP_2025} and SePiCo~\cite{SePiCo_2023} optimize for rare categories to boost mIoU, they often suffer from severe degradation on essential navigable regions. For instance, these methods exhibit poor performance on critical categories like road and sidewalk, dropping to approximately 77$\%$ and 37$\%$ IoU, respectively. Conversely, SPR+ST maintains superior performance on these major classes, reaching 97.9$\%$ IoU on road and 52.6$\%$ IoU on sidewalk. 
These significant performance gaps, exceeding 20$\%$ and 15$\%$ respectively, validate that structural regularization effectively maintains the global semantic layout required for safe autonomous driving.
Finally, Fig.~\ref{fig.synthia_pic} presents qualitative results for selected samples, visually confirming the segmentation quality of SPR.

\begin{figure*}[t!]
    \centering
    \includegraphics[width=0.99\textwidth]{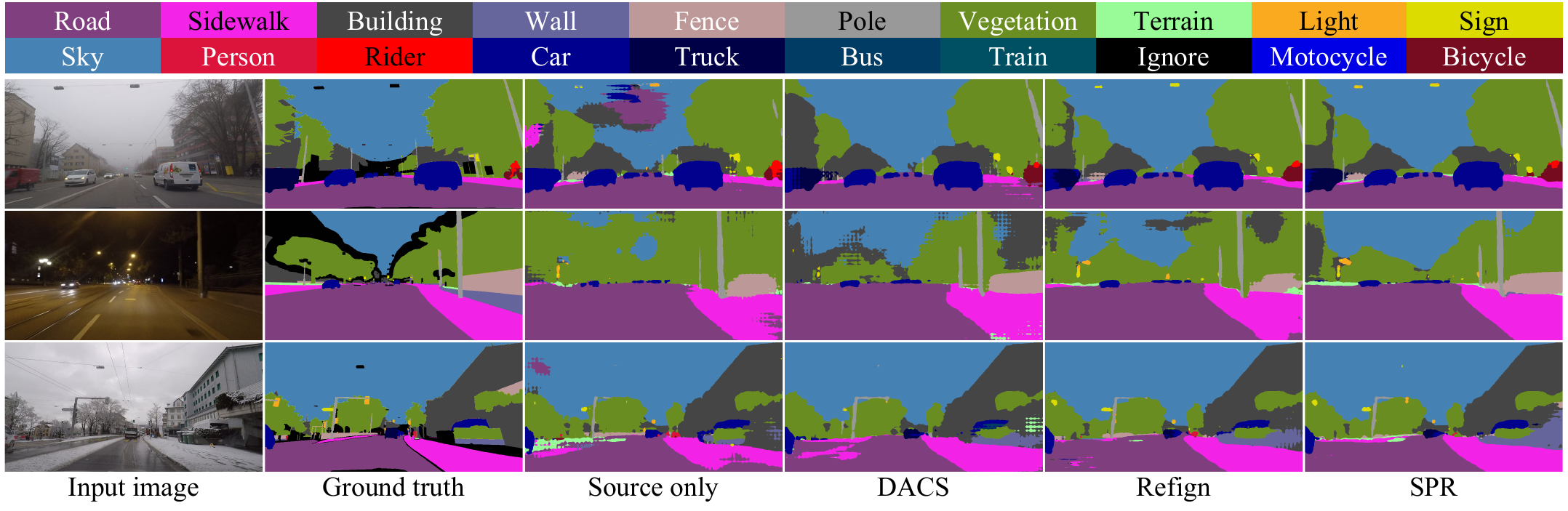}
    \caption{Qualitative results obtained on the Cityscapes $\to$ ACDC adaptation task. From left to right, the figure displays the input target-domain image and its corresponding ground truth, followed by the segmentation results predicted by four different methods: the source-only baseline model, DACS~\cite{Tranheden2021}, Refign~\cite{Refign}, and SPR.}
    \label{fig.acdc_pic}
\end{figure*}

\begin{table*}[t!]
	\centering
	\setlength{\heavyrulewidth}{2pt}
	\vspace{0.5em}
    \setlength{\tabcolsep}{0.6mm} 
    \tablesmallfont
	\caption{Quantitative comparison of state-of-the-art unsupervised domain adaptation methods for Cityscapes $\to$ ACDC adaptation task based on per-class IoU ($\%$) and overall mIoU ($\%$).}
	\begin{tabular}{lc|ccccccccccccccccccc|cc}
		\toprule[0.9pt]
		Method &Publication & \rotatebox[origin=c]{90}{Road} & \rotatebox[origin=c]{90}{Sidewalk} & \rotatebox[origin=c]{90}{Building} & \rotatebox[origin=c]{90}{Wall} & \rotatebox[origin=c]{90}{Fence} & \rotatebox[origin=c]{90}{Pole} & \rotatebox[origin=c]{90}{Light} & \rotatebox[origin=c]{90}{Sign} & \rotatebox[origin=c]{90}{Vegetation} & \rotatebox[origin=c]{90}{Terrain} & \rotatebox[origin=c]{90}{Sky} & \rotatebox[origin=c]{90}{Person} & \rotatebox[origin=c]{90}{Rider} & \rotatebox[origin=c]{90}{Car} & \rotatebox[origin=c]{90}{Truck} & \rotatebox[origin=c]{90}{Bus} & \rotatebox[origin=c]{90}{Train} & \rotatebox[origin=c]{90}{Motorbike}& \rotatebox[origin=c]{90}{Bike} & \rotatebox[origin=c]{90}{mIoU} & \rotatebox[origin=c]{90}{Gain} \\
		\midrule
        Source Only~\cite{Chen2017a} &TPAMI'17 &79.0 &21.8 &53.0 &13.3 &11.2 &22.5 &20.2 &22.1 &43.5 &10.4 &18.0 &37.4 &33.8 &64.1 &6.4 &0.0 &52.3 &30.4 &7.4 &28.8 &0.0 \\
        AdaptSeg~\cite{Tsai2018} &CVPR'18 &69.4 &34.0 &52.8 &13.5 &18.0 &4.3 &14.9 &9.7 &64.0 &23.1 &38.2 &38.6 &20.1 &59.3 &35.6 &30.6 &53.9 &19.8 &33.9 &33.4 &+4.6 \\
        ADVENT~\cite{Vu2019} &CVPR'19 &72.9 &14.3 &40.5 &16.6 &21.2 &9.3 &17.4 &21.2 &63.8 &23.8 &18.3 &32.6 &19.5 &69.5 &36.2 &34.5 &46.2 &26.9 &36.1 &32.7 &+3.9 \\
        BDL~\cite{Li2019} &CVPR'19 &56.0 &32.5 &68.1 &20.1 &17.4 &15.8 &30.2 &28.7 &59.9 &25.3 &37.7 &28.7 &25.5 &70.2 &39.6 &40.5 &52.7 &29.2 &38.4 &37.7 &+8.9 \\
        DANNet~\cite{DANNET} &CVPR'21 &\textbf{82.9} &53.1 &75.3 &32.1 &28.2 &26.5 &39.4 &40.3 &70.0 &39.7 &83.5 &42.8 &28.9 &68.0 &32.0 &31.6 &47.0 &21.5 &36.7 &46.3 &+17.5 \\
        DACS~\cite{Tranheden2021} &WACV'21 &58.5 &34.7 &76.4 &20.9 &22.6 &31.7 &32.7 &46.8 &58.7 &39.0 &36.3 &43.7 &20.5 &72.3 &39.6 &34.8 &51.1 &24.6 &38.2 &41.2 &+12.4 \\
        FDA~\cite{FDA} &ICLR'23 &73.2 &34.7 &59.0 &24.8 &\textbf{29.5} &28.6 &43.3 &44.9 &70.1 &28.2 &54.7 &47.0 &28.5 &74.6 &44.8 &\textbf{52.3} &63.3 &28.3 &39.5 &45.7 &+16.9 \\
        VBLC~\cite{vblc} &AAAI'23 &49.6 &39.3 &79.4 &35.8 &\textbf{29.5} &\textbf{42.6} &\textbf{57.2} &\textbf{57.5} &69.1 &\textbf{42.7} &39.8 &\textbf{54.5} &\textbf{29.3} &\textbf{77.8} &43.0 &36.2 &32.7 &\textbf{38.7} &\textbf{53.4} &47.8 &+19.0 \\
        Refign~\cite{Refign} &WACV'23 &49.5 &\textbf{56.7} &\textbf{79.8} &31.2 &25.7 &34.1 &48.0 &48.7 &\textbf{76.2} &42.5 &38.5 &48.3 &24.7 &75.3 &\textbf{46.5} &43.9 &\textbf{64.3} &34.1 &43.6 &48.0 &+19.2 \\
		\midrule
		\textbf{SPR (Ours)} &SCIS'26 	&77.8	&36.4	&78.9	&\textbf{36.1}	&29.1	&40.1	&46.9	&51.2	&74.9	&38.5	&\textbf{84.7}	&49.3	&22.3	&73.3	&40.4	&42.6	&57.0	&26.5	&34.5 &\textbf{49.5} &\textbf{+20.7} \\
        \bottomrule[0.9pt]
	\end{tabular}
	\label{tab.city2acdc}
\end{table*}

\subsubsection{Experimental results on Cityscapes → ACDC adaptation task}
\label{sec.exp.city_to_acdc}

To evaluate robustness against complex environmental variations in real-world scenarios, further experiments are conducted on the Cityscapes $\to$ ACDC adaptation task. Unlike the synthetic-to-real adaptation task, this setting involves severe visual degradation caused by adverse conditions such as fog, nighttime, rain, and snow.
The quantitative and qualitative experimental results are presented in Table~\ref{tab.city2acdc} and Fig.~\ref{fig.acdc_pic}.
The proposed SPR shows superior stability under these domain shifts. Specifically, the method achieves mIoU of 49.5$\%$, outperforming the source-only baseline by a substantial 20.7$\%$ and surpassing SoTA methods including VBLC~\cite{vblc} and Refign~\cite{Refign}.
A key advantage of SPR lies in its ability to preserve semantic structure when visual cues are compromised. For instance, on large-scale background categories such as the sky and wall, which are heavily affected by lighting and weather changes, SPR maintains remarkable consistency. Notably, the method achieves 84.7$\%$ IoU on the sky and 36.1$\%$ on the wall, significantly outperforming Refign, which achieves only 38.5$\%$ and 31.2$\%$, respectively. These results confirm that SPR effectively mitigates the impact of realistic texture corruption, ensuring reliable scene parsing performance even in the most challenging real-to-real adaptation tasks.

\subsection{Ablation study}
\label{sec.exp.eachpart}

\begin{table}[t!]
	\setlength{\heavyrulewidth}{2pt}
    \tablesmallfont
	\centering
	\setlength{\tabcolsep}{3.0mm}
	\caption{Ablation study of each component on GTA5 $\to$ Cityscapes adaptation task. ``P-Inter'' and ``P-Intra'' refer to prototypical inter-class and intra-class interactions, respectively; ``P-Pixel'' denotes prototype-pixel interaction; ``L'' represents pixel-wise attention for contrastive loss; and ``Ada-ST'' indicates adaptive threshold self-training.}
	\begin{tabular}
		{cccccc|c}
		\toprule[0.9pt]
        Source Only &P-Intra &P-Inter &P-Pixel &L &Ada-ST &mIoU (\%) \\
		\midrule
		\checkmark & & & & & &37.1 \\
  	    \checkmark &\checkmark & & & & &51.5 \\
        \checkmark & &\checkmark & & & &49.7 \\
        \checkmark &\checkmark &\checkmark & & & &52.2 \\
        \checkmark & & &\checkmark &\checkmark & &51.6 \\
        \checkmark &\checkmark &\checkmark &\checkmark & & &52.7 \\
        \checkmark &\checkmark &\checkmark &\checkmark &\checkmark & &53.1 \\
        \midrule
        \checkmark & & & & &\checkmark &45.5 \\
        \checkmark &\checkmark &\checkmark &\checkmark &\checkmark &\checkmark &\textbf{54.4} \\
        \bottomrule[0.9pt]
	\end{tabular}
	\label{tab.each_part}
\end{table}

\begin{figure*}[t!]
    \centering
    \includegraphics[width=0.99\textwidth]{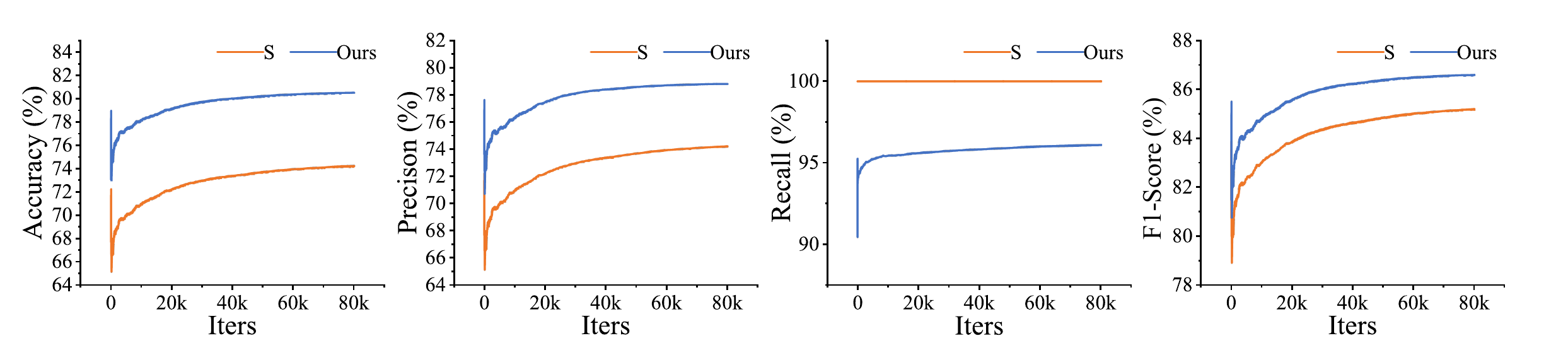}
    \caption{Performance comparison of pseudo label filtering strategies.
            ``S'' represents the softmax-based pseudo label filtering method with fixed probability thresholds.}
    \label{fig.pseudo_label_acc}
\end{figure*}

To evaluate the effectiveness of individual components, ablation experiments are conducted on the GTA5 $\to$ Cityscapes adaptation task, as detailed in Table~\ref{tab.each_part}. 
The source-only baseline, based on DeepLab-v2 with a ResNet-101 backbone, achieves mIoU of 37.1$\%$ on the Cityscapes validation set.
Specifically, the structural regularization mechanisms serve as the primary drivers of performance improvement. 
Independent applications of prototypical intra-class interaction, prototypical inter-class interaction, and prototype-pixel interaction each yield substantial gains of over 12$\%$, demonstrating that enforcing semantic consistency at both the prototype and pixel levels is critical for bridging the domain gap.
The integration of these complementary modules further refines feature alignment, establishing a robust foundation and achieving mIoU of 53.1$\%$.
Notably, the framework exhibits strong potential for further optimization via self-training. While applying self-training to the source-only baseline yields 45.5$\%$ mIoU, coupling it with the proposed SPR further boosts performance to 54.4$\%$. This improvement confirms that the structurally regularized feature distribution facilitates effective self-training on the target domain.

\begin{figure*}[t!]
    \centering
    \includegraphics[width=0.99\textwidth]{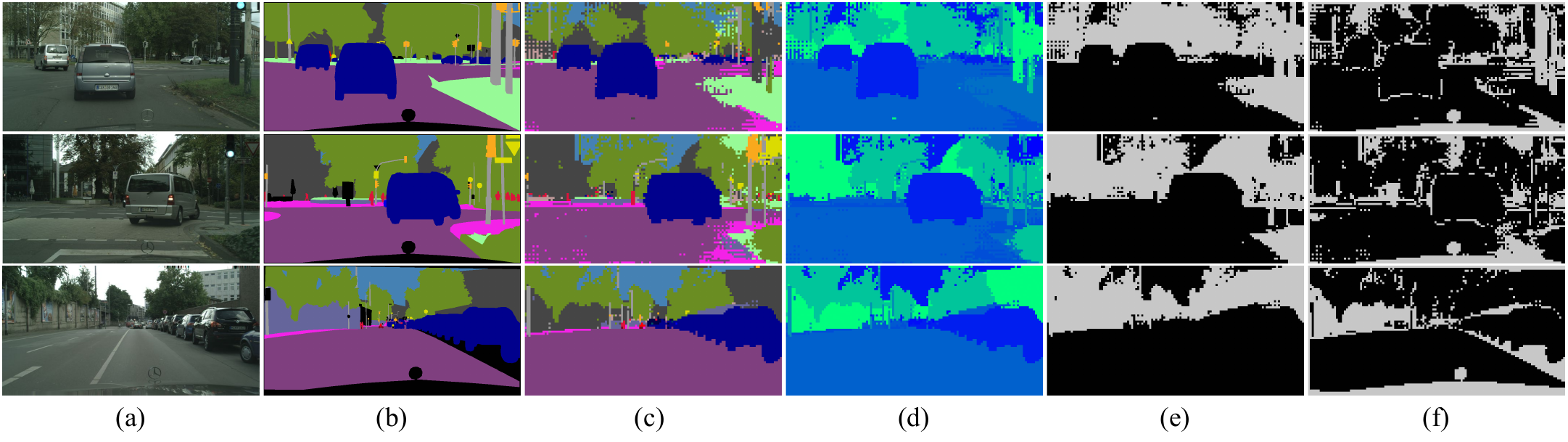}
    \caption{Visualization of pseudo label filtering. From left to right: (a) input target-domain images, (b) ground truth, (c) segmentation results of SPR, (d) entropy maps of SPR segmentation results, (e) mask for pseudo label filtering, and (f) ground truth mask for pseudo label filtering.}
    \label{fig.pseudo_label}
\end{figure*}

\begin{table}[t!]
	\setlength{\heavyrulewidth}{2pt}
    \tablesmallfont
	\centering
	\setlength{\tabcolsep}{2.5mm}
	\caption{Sensitivity analysis of the hyperparameter $\alpha$ for partitioning target-domain predictions into reliable and unreliable subsets for pseudo label filtering.}
	\begin{tabular}
		{c|c c c c c c c c c}
		\toprule[0.9pt]
        \multicolumn {10}{c}{GTA5 $\to$ Cityscapes} \\
		\midrule
		{$\alpha$} &0.1 &0.2 &0.3 &0.4 &0.5 &0.6 &0.7 &0.8 &0.9 \\
        \midrule 
		mIoU (\%) &51.8 &52.2 &52.1 &51.9 &51.0 &52.4 &52.5 &\textbf{54.4} &52.7\\
        \bottomrule[0.9pt]
	\end{tabular}
	\label{tab.alpha}
\end{table}

\subsection{Parameter sensitivity analysis}
\label{sec.exp.parameter_sensitivity}

\begin{figure*}[t!]
    \centering
    \includegraphics[width=0.99\textwidth]{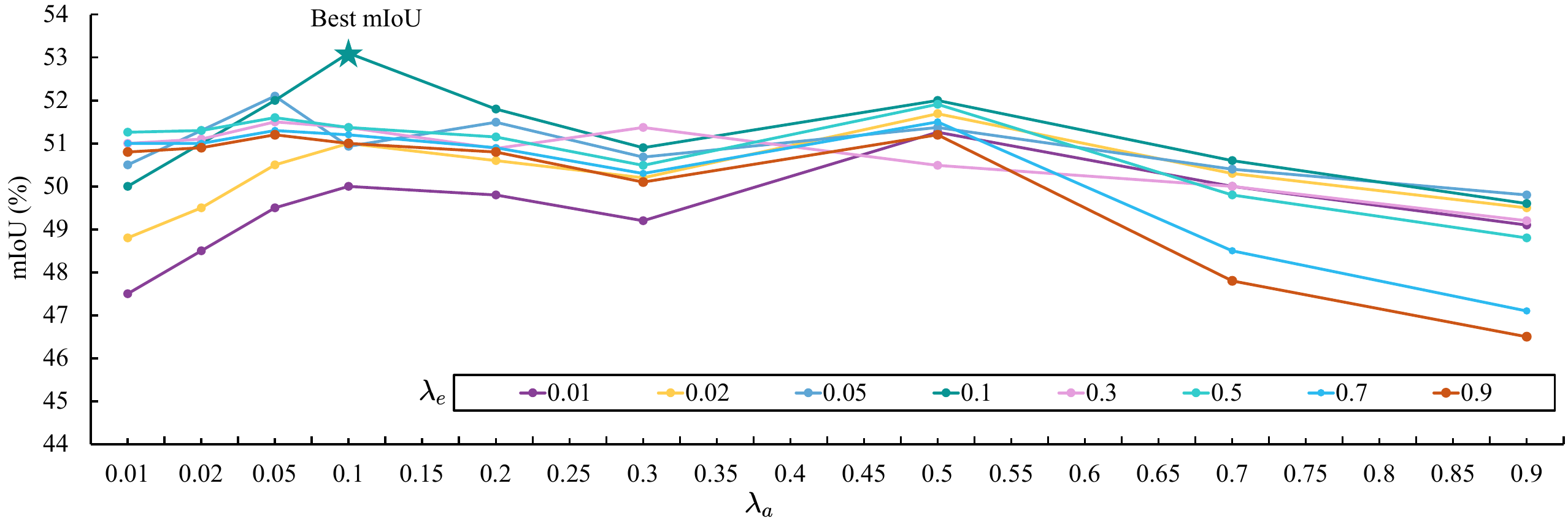}
    \caption{Analysis for the hyperparameters $\lambda_{e}$ and $\lambda_{a}$ of the proposed structured prototype regularization.}
    \label{fig.hyper_param}
\end{figure*}

The performance of the proposed SPR is influenced by two critical sets of hyperparameters: (1) the pseudo label generation threshold $\alpha$, which determines the selection of reliable pixels during the filtering process, and (2) the regularization weights $\lambda_{a}$ and $\lambda_{e}$, which modulate the intensity of intra-class and inter-class structural prototype regularization terms. A detailed analysis of their respective effects is presented in the following subsections.

\textbf{Pseudo label threshold}:
To validate the effectiveness of prototype-pixel interaction for pseudo label generation, a comprehensive evaluation is conducted. In contrast to conventional methods that rely on softmax logits~\cite{Jiang2022}, the proposed SPR leverages similarity between features and prototypes to filter high-confidence pixels. This selection process is controlled by the threshold parameter $\alpha$, which determines the proportion of reliable pixels retained. As evidenced by the sensitivity analysis in Table~\ref{tab.alpha}, optimal mIoU of 54.4$\%$ is achieved when $\alpha$ is set to 0.8, representing the most effective balance between label quantity and quality.
Furthermore, the quality of generated pseudo labels is evaluated against ground truth labels using standard classification metrics. 
As illustrated in Fig.~\ref{fig.pseudo_label_acc}, the comparison with the logit-based baseline~\cite{Jiang2022} reveals a distinct performance trade-off.
While the reference method exhibits higher recall, this suggests the inclusion of noisier labels. In contrast, the proposed SPR significantly enhances precision and accuracy, ultimately surpassing the baseline in F1-score. 
This result indicates that a cleaner and more accurate label set contributes more effectively to adaptation than a larger yet noisier one, further suggesting the superior capability of SPR in filtering ambiguous regions.
To qualitatively illustrate the effect of pseudo label filtering, Fig.~\ref{fig.pseudo_label} shows that the proposed strategy preserves reliable regions while suppressing uncertain predictions, producing high-confidence pseudo labels closely aligned with the ground truth masks.

\textbf{Structural regularization weights}:
The impact of intra-class and inter-class regularization weights, $\lambda_{a}$ and $\lambda_{e}$, is further analyzed on the GTA5 $\to$ Cityscapes adaptation task, as illustrated in Fig.~\ref{fig.hyper_param}.
Optimal performance is achieved with moderate weights ($\lambda_{a} = \lambda_{e} = 0.1$). This configuration provides balanced structural constraints that enhance both inter-class separability and intra-class compactness, while preserving the discriminative capacity of the prototypes. 
Consequently, distinct class boundaries are maintained while allowing sufficient flexibility for target-specific variations, thereby ensuring reliable feature alignment.
In scenarios with minimal regularization ($\lambda_{a} = \lambda_{e} = 0.01$), structural guidance becomes negligible. 
Without sufficient regularization, the prototypes act merely as source-domain class centroids, making the adaptation process equivalent to a standard prototype-based approach.
This functional similarity results in performance comparable to that of the baseline~\cite{Jiang2022}, thereby underscoring the critical role of the proposed SPR.
Conversely, excessive weights ($\lambda_{a} = \lambda_{e} = 0.9$) impose overly rigid constraints on the prototypes. Such over-regularization restricts representational capacity and degrades adaptation performance, highlighting the necessity of balanced regularization.
These results indicate that moderate structural constraints optimize adaptation by balancing inter-class separability and intra-class compactness effectively. Furthermore, the stable performance observed across a broad intermediate range ($0.05$–$0.3$) demonstrates the framework's robustness to hyperparameter variations.

\subsection{Computational costs and runtime performance}
\label{sec.exp.Computational_Costs}

\begin{table}[!t]
\caption{Comparison of training efficiency and adaptation performance for different unsupervised domain adaptation methods.}
\label{table.runtime_train}
\tablesmallfont
\setlength{\tabcolsep}{5.8mm}
\centering
\begin{tabular}{l|cccc}
\toprule
\textbf{Method} &{\textbf{Epoch} \textbf{Duration (s)} $\downarrow$} & \textbf{Each Iter (s) $\downarrow$} & \textbf{Memory (MB) $\downarrow$} & \textbf{mIoU (\%) $\uparrow$} \\
\midrule
AdaptSeg~\cite{Tsai2018} &1561 &0.525 &11251 &42.4 \\
UIDA~\cite{Pan2020} &1560 &0.525 &10277 &46.3 \\
FADA~\cite{Wang2020a} &2182 &0.734 &\textbf{5218} &49.2 \\
ASA~\cite{Zhou2020} &1570 &0.520 &12855 &45.1 \\
DACS~\cite{Tranheden2021} &\textbf{1147} &\textbf{0.380} &11483 &52.1 \\
MetaCorrect~\cite{Guo2021} &15358 &5.163 &19679 &52.1 \\
DAST~\cite{Yu2021} &2181 &0.733 &11685 &49.6 \\
ProDA~\cite{Zhang2021} &2582 &0.868 &11469 &53.7 \\
ProCA~\cite{Jiang2022} &1254 &0.422 &21640 &48.8 \\
\midrule
\textbf{SPR (Ours)} &1332 &0.448 &17825 &\textbf{54.4} \\
\bottomrule
\end{tabular}
\end{table}

\begin{table}[!t]
\caption{Comparison of inference efficiency and segmentation performance for different segmentation backbones.}
\label{table.runtime_test}
\tablesmallfont
\setlength{\tabcolsep}{2.1mm}
\centering
\begin{tabular}{c|cccccccc}
\toprule
\textbf{Family} & \textbf{Backbone} & \textbf{Params (M) $\downarrow$} & \textbf{FLOPs (G) $\downarrow$} & 
\textbf{FPS $\uparrow$} & \textbf{Latency (ms) $\downarrow$} & \textbf{Memory (MB) $\downarrow$} & \textbf{mIoU (\%) $\uparrow$} \\
\midrule
\multirow{5}{*}{\centering \textbf{ResNet~\cite{He2016}}} 

 & ResNet-18  & \textbf{11.5}  & 101.6 & 82.0  & 12.2  & \textbf{144} & 35.6 \\
 & ResNet-34  & 21.6  & 184.7 & 45.5  & 22.0  & 260 & 38.3 \\
 & ResNet-50  & 24.9  & 211.9 & 39.4  & 25.3  & 298 & 46.4 \\
 & ResNet-101 & 43.9  & 367.9 & 23.0  & 43.4  & 518 & 54.4 \\
 & ResNet-152 & 59.5  & 496.4 & 17.0  & 58.9  & 698 & \textbf{55.0} \\

\midrule
\multirow{4}{*}{\centering \textbf{VGG~\cite{vgg}}} 
 & VGG11 &19.6 &\textbf{77.8}  &\textbf{120.9} &\textbf{8.2}  &321 &37.3 \\
 & VGG13 &23.3 &112.5 &83.9  &11.9 &354 &39.7 \\
 & VGG16 &29.5 &190.7 &60.7  &16.4 &428 &46.5 \\
 & VGG19 &34.8 &234.2 &51.0  &19.5 &436 &48.6 \\
 
\bottomrule
\end{tabular}
\end{table}

\begin{figure*}[t!]
    \centering
    \includegraphics[width=0.99\textwidth]{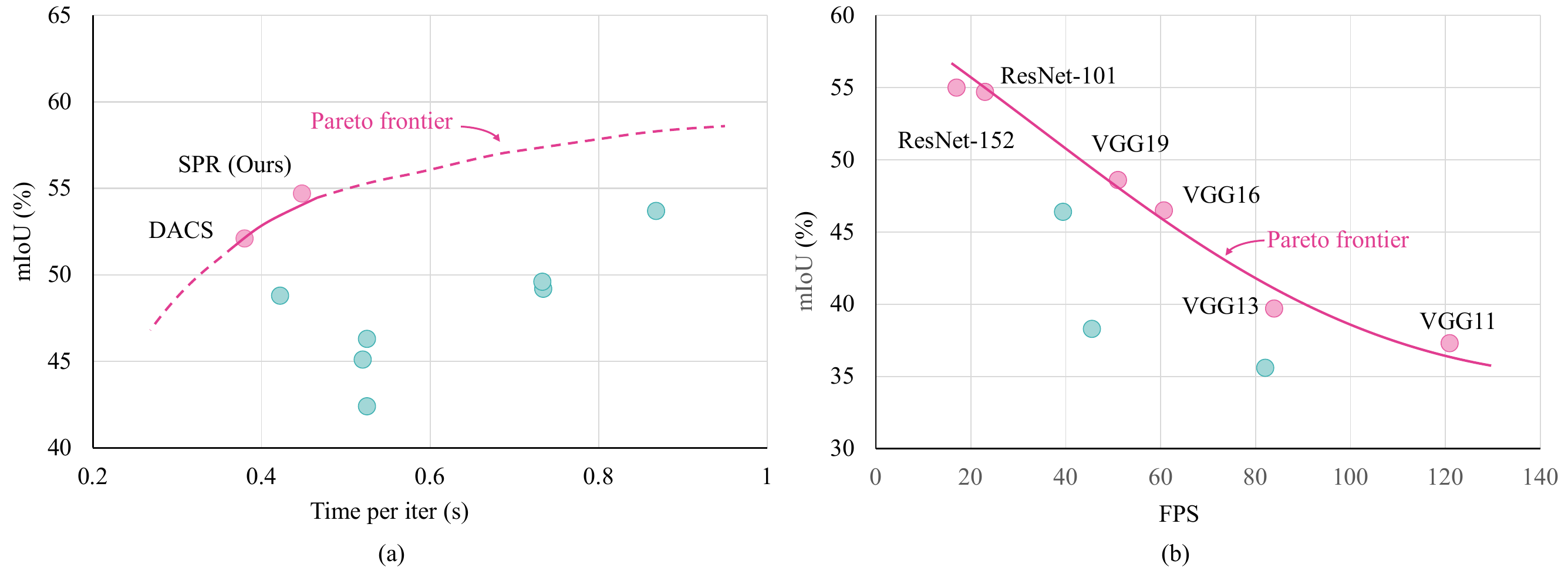}
    \caption{Analysis of the trade-off between performance and runtime. (a) Performance and training efficiency across UDA methods. (b) Performance and inference efficiency across segmentation backbones. The Pareto-frontier highlights non-dominated methods. The proposed method lies on the frontier and exhibits a favorable trade-off.}
    \label{fig.pf_figure}
\end{figure*}

Computational efficiency is crucial for driving scene parsing models, particularly in real-world autonomous driving scenarios. To provide a comprehensive evaluation, computational costs and runtime performance are analyzed for both training and inference phases.
Experiments are conducted on the GTA5~\cite{Richter2016} $\to$ Cityscapes~\cite{Cordts2016} adaptation task. The training process is restricted to 20 epochs, comprising 2,975 images. The batch size is set to 1, with input images resized to $512\times1,024$ pixels.
Table~\ref{table.runtime_train} details the epoch duration, iteration time, memory usage, and mIoU for representative UDA methods. Specifically, compared with ProDA~\cite{Zhang2021}, SPR reduces the training time by nearly 50$\%$ while achieving a 0.7$\%$ improvement in mIoU. Although it is slightly slower than DACS~\cite{Tranheden2021}, SPR yields a substantial performance gain of 2.3$\%$. 
These results indicate that the proposed SPR method introduces only moderate training complexity relative to other SoTA methods implemented on the same driving scene parsing model. 
Consequently, SPR achieves a superior trade-off between computational efficiency and segmentation accuracy, which is advantageous for practical deployment in autonomous driving systems.

During the inference phase, the computational complexity is primarily determined by the backbone architecture, as the proposed SPR method does not introduce any additional parameters or modules into the inference pipeline. Table~\ref{table.runtime_test} summarizes the FPS, latency, memory usage, and mIoU across representative backbones. These metrics serve as key indicators of deployment feasibility.
The results demonstrate that SPR maintains a favorable balance between accuracy and efficiency.
For example, with a ResNet-101 backbone, SPR achieves 54.4$\%$ mIoU, 23 FPS, and 43.4 ms latency. 
These metrics confirm the capability of the proposed SPR method for real-time application. Consequently, this demonstrates that SPR significantly enhances segmentation accuracy without compromising inference efficiency, thereby effectively balancing performance improvements and computational overhead.
Furthermore, adopting lightweight backbones such as ResNet-18 and VGG11 significantly increases inference speed with only a moderate drop in accuracy, offering flexible solutions for resource-constrained driving scenarios.

Based on the quantitative results reported in Table~\ref{table.runtime_train} and Table~\ref{table.runtime_test}, Fig.~\ref{fig.pf_figure} further visualizes the trade-off between segmentation performance and computational efficiency during both training and inference. As shown, the proposed SPR lies on the Pareto frontier, indicating that it achieves a competitive balance between accuracy and runtime without being dominated by alternative methods.

\section{Discussion}
\label{sec.discussion}

\subsection{Theoretical analysis}

Domain shift arises from the discrepancy between the joint probability distributions over images and labels in the source and target domains, \ie $P^s(\mathcal{X}^s,\mathcal{Y}^s) \neq P^t(\mathcal{X}^t,\mathcal{Y}^t)$, resulting in degraded performance when a source-domain scene parsing model is directly applied to the target domain. 
Specifically, UDA aims to learn a hypothesis $h$ that minimizes the target-domain prediction error $\epsilon^t(h)$, even in the absence of target-domain annotations. 
Conventionally, this is achieved by aligning source- and target-domain features or model outputs so that the probability distributions, $P^s(h(\mathcal{X}^s) \mid \mathcal{X}^s)$ and $P^t(h(\mathcal{X}^t) \mid \mathcal{X}^t)$, become similar across domains. 
Conventional UDA methods mainly reduce the marginal gap, \ie $P^s(\mathcal{X}^s) \approx P^t(\mathcal{X}^t)$, but do not ensure the alignment of conditional probability distributions $P^s(\mathcal{Y}^s \mid \mathcal{X}^s)$ and $P^t(\mathcal{Y}^t \mid\mathcal{X}^t)$, which often results in high intra-class variability and low inter-class separability.
The proposed SPR method mitigates this by employing structured regularization to explicitly preserve the structure of class-specific features in the target domain, which effectively promotes inter-class separability and intra-class compactness.
Let class-specific prototypes be denoted as $\boldsymbol{p}^s_{c}$ and $\boldsymbol{p}^t_{c}$, and the method enforces:
\begin{equation}
Corr(\{\boldsymbol{p}^s_{c}\}_{c=1}^C) \approx Corr(\{\boldsymbol{p}^t_{c}\}_{c=1}^C),
\end{equation}
where $Corr(\cdot)$ denotes the calculation of the correlation matrix. This preserves inter-class separability and intra-class compactness, guiding $P^t(\mathcal{Y}^t \mid\mathcal{X}^t)$ to approximate $P^s(\mathcal{Y}^s \mid \mathcal{X}^s)$ in a class-aware manner.
Leveraging the theoretical framework for domain adaptation  by Ben-David~\emph{et al.}~\cite{ben2010theory},
the expected prediction error on the target domain for a hypothesis $h$ is bounded as follows:
\begin{equation}
\epsilon^t(h) \le \epsilon^s(h) + \frac{1}{2} d(P^s, P^t) + \lambda,
\end{equation}
where $\epsilon^s(h)$ denotes the expected prediction error of $h$ on the source domain, which is accessible during training, and $\epsilon^t(h)$ denotes the expected prediction error on the target domain, which remains unknown due to the lack of target-domain annotations. The term $d(P^s, P^t)$ quantifies the divergence between the source- and target-domain marginal feature distributions, while $\lambda$ quantifies the error arising from misaligned class-conditional distributions $P^s(\mathcal{Y}^s \mid \mathcal{X}^s)$ and $P^t(\mathcal{Y}^t \mid\mathcal{X}^t)$. Conventional UDA methods mainly focus on reducing $d(P^s, P^t)$ but have limited control over $\lambda$, which tends to remain substantial when intra-class compactness and inter-class separability are compromised. 
By enhancing intra-class compactness and enforcing consistent inter-class correlations across domains, SPR effectively reduces $\lambda$ while maintaining a low $d(P^s, P^t)$, thereby tightening the theoretical bound on $\epsilon^t(h)$ and enabling more reliable adaptation.

\subsection{Limitations and prospective extensions}

\subsubsection{Computational complexity and efficiency trade-offs}

Experiments demonstrate that the proposed method maintains computational efficiency. In terms of training cost, its performance is comparable to existing UDA methods.
Notably, the domain adaptation strategy incurs no additional computational overhead during inference. This efficiency enables the final model to satisfy deployment requirements in real time.
However, the prototype interaction mechanism, which computes correlations between matrices, results in a slight increase in training time compared with conventional methods.
Consequently, strategies to simplify these computations are explored, and additional experiments are conducted to evaluate their impact on both performance and efficiency.
Specifically, a decoupling strategy is adopted to mitigate the computational overhead associated with calculating and storing multiple correlation matrices.
The correlation matrices for classes and channels, $\boldsymbol{R}_{e}^l \in \mathbb{R}^{C \times C}$ and $\boldsymbol{R}_{a}^l \in \mathbb{R}^{D \times D}$, are derived directly from the global class-specific prototypes $\boldsymbol{P} \in \mathbb{R}^{D \times C}$.
This modification reduces the space complexity from $\mathcal{O}(D \cdot C^2 + C \cdot D^2)$ to $\mathcal{O}(C^2 + D^2)$, eliminating redundant computations in high dimensions. 

\begin{table}[!t]
\caption{Comparison of training efficiency and adaptation performance for the original and simplified SPR framework.}
\label{table.Simplified}
\tablesmallfont
\setlength{\tabcolsep}{3.5mm}
\centering
\begin{tabular}{c|cccc}
\toprule
\textbf{Method} &{\textbf{Epoch Duration (s)} $\downarrow$} & \textbf{Each Iter (s) $\downarrow$} & \textbf{Memory (MB) $\downarrow$} & \textbf{mIoU (\%) $\uparrow$} \\
\midrule
SPR (Simplified) &1025 &0.344 &13536 &{47.9} \\
SPR (Original) &1332 &0.448 &17825 &{54.4} \\
\bottomrule
\end{tabular}
\end{table}

Comparative results are presented in Table~\ref{table.Simplified}. While the simplified version effectively reduces memory usage, a performance gap in mIoU is observed when compared to the original design. 
This performance gap can be attributed to two structural distinctions: (1) The simplified method relies on a single global correlation matrix shared across all classes, whereas distinct semantic classes exhibit unique feature dependencies. In contrast, the proposed SPR preserves interaction matrices for individual classes, enabling the model to capture unique structural correlations for each category. (2) The results indicate that the higher dimensionality of the proposed SPR is not redundant; rather, it plays a critical role in preserving fine-grained semantic structure and ensuring superior performance.
Current observations confirm that the interaction in high dimensions is essential for optimal performance. Nevertheless, there is still potential for further optimization of the proposed method. Consequently, future research will explore more efficient approximation techniques to reduce memory usage during training, with the objective of maintaining robust correlation modeling for class-specific features.

\begin{table*}[t!]
	\centering
	\setlength{\heavyrulewidth}{2pt}
	\vspace{0.5em} 
    \setlength{\tabcolsep}{0.7mm} 
    \tablesmallfont
	\caption{Quantitative comparison with state-of-the-art source-free unsupervised domain adaptation methods on the GTA5 $\to$ Cityscapes adaptation task based on per-class IoU ($\%$) and overall mIoU ($\%$).}
	\begin{tabular}{lc|c|ccccccccccccccccccc|c}
		\toprule[0.9pt]
		Method &Publication &\rotatebox[origin=c]{90}{Source-Free} & \rotatebox[origin=c]{90}{Road} & \rotatebox[origin=c]{90}{Sidewalk} & \rotatebox[origin=c]{90}{Building} & \rotatebox[origin=c]{90}{Wall} & \rotatebox[origin=c]{90}{Fence} & \rotatebox[origin=c]{90}{Pole} & \rotatebox[origin=c]{90}{Light} & \rotatebox[origin=c]{90}{Sign} & \rotatebox[origin=c]{90}{Vegetation} & \rotatebox[origin=c]{90}{Terrain} & \rotatebox[origin=c]{90}{Sky} & \rotatebox[origin=c]{90}{Person} & \rotatebox[origin=c]{90}{Rider} & \rotatebox[origin=c]{90}{Car} & \rotatebox[origin=c]{90}{Truck} & \rotatebox[origin=c]{90}{Bus} & \rotatebox[origin=c]{90}{Train} & \rotatebox[origin=c]{90}{Motorbike}& \rotatebox[origin=c]{90}{Bike} & \rotatebox[origin=c]{90}{mIoU} \\
        
        \midrule
        SFDA~\cite{Liu_2021_CVPR} &CVPR'21 &\Checkmark &84.2 &39.2 &82.7 &27.5 &22.1 &25.9 &31.1 &21.9 &82.4 &30.5 &85.3 &58.7 &22.1 &80.0 &33.1 &31.5 &3.6 &27.8 &30.6 &43.2 \\

        SFUDA~\cite{ye2021source} &ACM MM'21 &\Checkmark &\textbf{95.2} &40.6 &85.2 &30.6 &26.1 &35.8 &34.7 &32.8 &85.3 &41.7 &79.5 &61.0 &28.2 &\textbf{86.5} &41.2 &45.3 &\textbf{15.6} &33.1 &40.0 &49.4 \\

        SimT~\cite{Guo_2022_CVPR} &CVPR'22 &\Checkmark &88.5 &45.4 &85.2 &38.4 &28.0 &34.9 &40.8 &37.0 &84.9 &42.4 &80.4 &60.6 &26.2 &85.6 &38.0 &48.0 &0.0 &35.1 &35.9 &49.2 \\

        CSFDA~\cite{Karim_2023_CVPR} &CVPR'23 &\Checkmark &90.4 &42.2 &83.2 &34.0 &29.3 &34.5 &36.1 &38.4 &84.0 &43.0 &75.6 &60.2 &28.4 &85.2 &33.1 &46.4 &3.5 &28.2 &44.8 &48.3 \\
        
        ATP~\cite{wang2024curriculum} &T-PAMI'24  &\Checkmark &93.2 &\textbf{55.8} &{86.5} &45.2 &27.3 &36.6 &42.8 &37.9 &86.0 &43.1 &\textbf{87.9} &\textbf{63.5} &15.3 &85.5 &41.2 &{55.7} &0.0 &38.1 &\textbf{57.4} &52.6 \\

        DBC~\cite{Tiandbc} &T-MM'24 &\Checkmark &89.3 &38.7 &85.7 &34.1 &28.5 &\textbf{39.1} &\textbf{43.5} &\textbf{44.4} &86.2 &36.0 &84.5 &60.2 &25.2 &84.0 &38.2 &49.5 &6.5 &32.2 &45.9 &50.1\\

        IAPC~\cite{Caoiapc} &T-IV'24  &\Checkmark &90.9 &36.5 &84.4 &36.1 &\textbf{31.3} &32.9 &39.9 &38.7 &84.3 &38.6 &87.5 &58.6 &28.8 &84.3 &33.8 &49.5 &0.0 &34.1 &47.6 &49.4\\

        \midrule
        \textbf{SPR (Ours)} &SCIS'26 &\Checkmark &{86.0} &25.6 &81.9 &39.5 &23.2 &19.7 &29.6 &17.3 &{83.9} &{36.0} &{78.9} &57.6 &{31.2} &75.6 &31.9 &37.2 &0.2 &26.2 &{42.6} &43.4 \\

        \textbf{SPR (Ours)} &SCIS'26 &\XSolidBrush &90.4 &40.7 &\textbf{91.5} &\textbf{47.1} &30.8 &32.8 &41.0 &33.2 &\textbf{91.1} &\textbf{58.9} &87.7 &61.3 &\textbf{35.0} &79.1 &\textbf{51.5} &\textbf{65.6} &0.1 &\textbf{43.7} &52.6 &\textbf{54.4} \\
        
        \bottomrule[0.9pt]
	\end{tabular}
	\label{tab.sfuda}
\end{table*}

\subsubsection{Dependency on source-domain annotations and comparison with source-free UDA methods}

The primary objective of this study is to address challenges in domain adaptation from synthetic to real environments, where labeled source-domain data are available during training.
To further assess the impact of this dependency, a comprehensive comparison with representative source-free unsupervised domain adaptation (SFUDA) methods is conducted on the GTA5 $\to$ Cityscapes adaptation task. 
As presented in Table~\ref{tab.sfuda}, the proposed SPR leverages a labeled dataset from the source domain to achieve superior performance compared to representative SFUDA methods.
Additionally, to simulate a source-free scenario, supervision from source-domain data is disabled within the framework. Although SPR is not specifically developed for the SFUDA setting, it achieves performance comparable to several representative SFUDA methods.
However, a noticeable performance gap persists when compared to leading SFUDA methods. This indicates that the naive removal of supervision from the source-domain data is insufficient to preserve structural consistency. 
Consequently, extending SPR to settings without source-domain data would require the integration of self-supervised objectives to compensate for the absence of explicit cross-domain supervision. 
Accordingly, investigating efficient self-training strategies to bridge this gap remains a priority for future research.
The proposed SPR is presented as a viable framework for driving scene parsing in domain adaptation from synthetic to real environments, offering a favorable balance between performance and efficiency.

\section{Conclusion and future work}
\label{sec.conclusion}

In this article, a novel UDA framework was proposed for driving scene parsing in synthetic-to-real scenarios, which leveraged contrastive learning to explicitly model the structural relations within the semantic feature space. The proposed method constructed and regularized class prototypes and their correlation matrices, effectively promoting a well-structured semantic feature space and thereby facilitating more precise domain alignment.
Specifically, the proposed approach introduced three key improvements: (1) explicit quantification and regularization of inter-class and intra-class relations within the feature space, (2) integration of an entropy-based noise filtering mechanism to enhance the quality of pseudo labels in the target domain, and (3) refinement of pixel-level feature alignment through a pixel-level attention mechanism embedded in the contrastive learning process.
Extensive experiments on multiple driving scene parsing benchmarks demonstrated the effectiveness of the proposed method compared to state-of-the-art UDA techniques.
Future work will involve extending the proposed framework to the source-free setting, where source data is inaccessible for prototype initialization. To address this constraint, strategies for adapting structural regularization and alignment mechanisms will be explored, potentially by leveraging target-domain statistics or self-supervised learning.

\section{Acknowledgment}
\label{sec.ack}
This research was supported by the National Natural Science Foundation of China under Grants 62473288, 62388101, 62333017, and 62233013, the Fundamental Research Funds for the Central Universities, and the Xiaomi Young Talents Program.

\end{document}